\documentclass{article}

\usepackage[nonatbib,final]{neurips_2022} 

\usepackage[utf8]{inputenc} 
\usepackage[T1]{fontenc}    
\usepackage{url}            
\usepackage{amsfonts}       
\usepackage{nicefrac}       
\usepackage{microtype}      
\usepackage[dvipsnames]{xcolor}         
\usepackage{subcaption}
\usepackage{graphicx}
\usepackage{enumitem}
\usepackage{amsmath}
\usepackage{amssymb}
\usepackage{booktabs}
\usepackage{comment}
\usepackage{array}
\usepackage{makecell}
\usepackage{multirow}
\usepackage{enumitem}
\usepackage{algorithm}
\usepackage[noend]{algpseudocode}
\usepackage{comment}
\setlist{nolistsep}
\usepackage{nicematrix}
\usepackage{multirow}
\usepackage{dsfont}
\usepackage{wrapfig}
\usepackage{mathtools}
\usepackage{subcaption}
\newcolumntype{M}[1]{>{\centering\arraybackslash}m{#1}}
\newcolumntype{N}{@{}m{0pt}@{}}

\usepackage{hyperref}

\newcommand*{\nolink}[1]{  \begin{NoHyper}#1\end{NoHyper} }

\usepackage[capitalize]{cleveref}

\begin{document}

\title{Towards Disentangling Information Paths with \\ Coded ResNeXt}


\author{Apostolos Avranas\thanks{Currently working at Amadeus, Nice}\\ EURECOM\\ Sophia Antipolis, France\\ \texttt{avranas@eurecom.fr}\\ \And
Marios Kountouris\\ EURECOM \\ Sophia Antipolis, France \\ \texttt{kountour@eurecom.fr}}

\maketitle

\begin{abstract}
The conventional, widely used treatment of deep learning models as black boxes provides limited or no insights into the mechanisms that guide neural network decisions. Significant research effort has been dedicated to building interpretable models to address this issue. Most efforts either focus on the high-level features associated with the last layers, or attempt to interpret the output of a single layer. In this paper, we take a novel approach to enhance the transparency of the function of the whole network. We propose a neural network architecture for classification, in which the information that is relevant to each class flows through specific paths. These paths are designed in advance before training leveraging coding theory and without depending on the semantic similarities between classes. A key property is that each path can be used as an autonomous single-purpose model. This enables us to obtain, without any additional training and for any class, a lightweight binary classifier that has at least $60\%$ fewer parameters than the original network. Furthermore, our coding theory based approach allows the neural network to make early predictions at intermediate layers during inference, without requiring its full evaluation. Remarkably, the proposed architecture provides all the aforementioned properties while improving the overall accuracy. We demonstrate these properties on a slightly modified ResNeXt model tested on CIFAR-10/100 and ImageNet-1k.
\end{abstract}

\section{Introduction}
\label{sec:intro}
Most successful deep learning architectures for image classification consist of a certain building block applied sequentially several times: one block follows another until a linear operation finally outputs the model prediction. In deep convolutional neural networks (CNNs), the block consists of multiple convolutional operations \cite{Conv1_lecun,Conv2_lecun} applied sequentially. Nonetheless, there are numerous proposals placing the convolutional layers in parallel, forming multi-branch designs. For instance, inception models \cite{InceptionV1,InceptionV3_labelSmoothing} use blocks with multiple branches, each applying some convolutional operations on the block's input and finally concatenating at the end the output of all branches. The multi-branch design framework can also accommodate skip connections \cite{ResNets}, as initially done in ResNeXt networks \cite{ResNeXt}, and later refined using squeeze-excitation in \cite{SqueezeExcitation}, or a split-attention mechanism in \cite{ResNest_versionOfResNeXt}. 
A first question we ask in this work is: \emph{what is the purpose of multi-branch architectures?}
    
Initially, in AlexNet \cite{AlexNet} two branches were employed to allow 
the distribution of the model across two GPUs, which at that time had limited memory. Nowadays, multi-branch architectures are commonly used for distributing the parameters of a block into 
branches such that each one applies a separate transformation to the input. However, rare are the cases where each branch is shown to contribute in a different way. One example is SKNet \cite{SelectiveKernel_versionOfResNeXt}, in which each branch is associated with different receptive field size, and zooming in or out of an input image activates the appropriate branch. Nonetheless, the value of multi-branch networks is mostly justified by achieving a higher accuracy.
Multi-branch blocks are also used for network architecture search, where the block/cell architecture is optimized selecting the number of branches, the operation that each performs, and how they are combined \cite{NAS_1,NAS_5,NAS_2,NAS_4,NAS_3}. Still, the focus therein was on improving accuracy.
    
In this work, we investigate how to ensure that each branch provably contributes in a different way in a multi-branch architecture. We propose a novel way to organize in a class-wise manner the transformations carried out by the branches. Before the training starts and without using any information on the semantic properties of the classes, we assign each branch to a specific set of classes. This set remains fixed throughout the training and the branches are trained to activate only to classes within that set. This behavior is achieved mainly by applying a loss function that pushes the output of the branches to be zero for the samples that do not belong to their assigned set of classes. 
Thanks to this assignment, once the network is trained, for any given class there is a unique path traversing the network through which the information related to that class flows. Conditioning on a class then, extracting only the parameters that participate in its unique path results in a model that has $60\%$ less parameters than the original one and operates as a binary classifier for that class. To showcase the unique features and the advantages of our idea, we use the state-of-the-art multi-branch architecture ResNeXt \cite{ResNeXt}, to which we perform a small number of modifications.
    
Our main contributions can be summarized as follows:
\begin{itemize}[leftmargin=*]
\item We modify the ResNeXt architecture so that it functions in a more transparent way by forcing the information related to each class to flow through well-defined network paths. 
\item As a proof of concept, we show that in order to form those paths, it is not necessary to rely on the semantics of each class and the similarity between classes.     
\item Without any additional training, we can obtain a single-purpose model per class operating as a binary classifier, which has $60\%$ fewer parameters as compared to the complete network.
\item We demonstrate that the intermediate layers can be used both for making early predictions and for providing a confidence level for correctness of the network's final prediction.
\item The proposed Coded ResNeXt significantly improves ResNeXt accuracy across all tested datasets.
\end{itemize}

\section{Related Work}
\label{sec:RelatedWork}

There have been numerous attempts to understand how deep neural networks actually work. For instance, activation maximization tries to find the input that increases neuron activation \cite{activationMax_1,activationMax_2,activationMax_3,activationMax_4}. 
Saliency maps \cite{saliencyMaps_1, saliencyMaps_2,saliencyMaps_3,saliencyMaps_4, saliencyMaps_5,saliencyMaps_6} find the pixels that have the largest influence on the model prediction. 
However, such approaches cannot really explain how the network operates, and they mainly serve as post hoc visualization methods \cite{StopExplainingBlackBox}. 
In contrast, there are many interpretable by design proposals\cite{ NeuralDecisionTrees,InterpretableCNN, Prototype1,Prototype2,ConceptBottleneckModels}; however, most of them focus on enhancing interpretability only in the last layers.
In \cite{NeuralDecisionTrees}, the final linear layer is replaced with a differentiable decision tree, and in \cite{InterpretableCNN}, a loss is used to make each filter of the very high-level convolutional layer represent a specific object part. In \cite{Prototype1}, the model's output is compared with learnt prototypes, whereas in \cite{ConceptBottleneckModels} represents concepts on which humans can intervene. 

Another direction for enhancing interpretability is through disentanglement. While there is not yet a generally accepted definition, disentanglement aims at separating the main factors that are present in the data distribution \cite{DisentanglementDefinition_v1,DisentanglementDefinition_v2,Disentanglement_Review}. Still, existing approaches focus on disentangling the factors at a single vector/tensor (the latent representation), which is either the input or the output of a network. 
This applies to (variational) autoencoders   \cite{Disentanglement_VAE_1,Disentanglement_VAE_2,Disentanglement_VAE_3,Disentanglement_VAE_4,Disentanglement_VAE_5}, generative adversarial networks \cite{Disentanglement_GAN_1,Disentanglement_GAN_2,Disentanglement_GAN_3}, normalizing flows  \cite{Disentanglement_NF_1,Disentanglement_NF_2}, or even architectures that aim to decompose the content to the style representation of an image \cite{Disentanglement_ContentSytle_1,Disentanglement_ContentSytle_2}. 
In contrast, we approach disentanglement as ``the way information travels through the network'' \cite{FundamentalPrinciples10Challenges} and look at the neural network architectures for classification
as a whole.
Specifically, our goal is to control the paths through which information flows 
by assigning each part of the network to a specific subsets of classes. 
Related to our work is \cite{Wang_2018}, where they interpret a deep neural network by identifying such information paths; one key difference is that they use a post hoc method, so the paths are identified and not designed. 

There are two additional lines of work related to ours. The first line proposes to dynamically control the path that a sample follows through the network 
by letting an extra network choose which parts to be pruned/omitted \cite{DynamicChannelRouting_1,DynamicChannelRouting_2,DynamicChannelRouting_3,DynamicChannelRouting_4,DynamicChannelRouting_5,DynamicChannelRouting_6,DynamicChannelRouting_7, DynamicChannelRouting_8,DynamicChannelRouting_9}.
This brings memory and speedup gains during inference, since the samples do not pass through the entire network. 
Our proposal has a few key differences with these works. 
First, we do not use any extra network that has to be trained to learn good paths. Second, our paths are defined prior to training (and thus not learnt).
Most importantly, these works do no guarantee that two samples 
of the same class follow the same path, 
so it is not possible to
extract class-specific network portions to use as
single-purpose models. 
The second line of work concerns models that can make early predictions without evaluating the whole network \cite{R4_Review_AddEarlyExits}. Similarly to some previous works \cite{R4_Early_Exits_initial,R4_OccamNets,Early_exit_IC_Threshold,Early_exit_IC_BERT,Early_exit_IC_Cascade_Ensemble}, we also apply loss functions to the middle layers which facilitates early predictions; however, all those works necessitate additional parameters that are trained as classifiers performing early predictions. In our case, since we force each class to have a unique activation footprint on every middle layer, the early predictions emerge naturally by just looking at the activation patterns created in the middle layers as the samples are forwarded.

While in this work we assign classes to parts of the network arbitrarily, an interesting possibility is to exploit semantic/visual similarities for the assignment, similarly to \cite{R1_Deng_HEX,R1_Deng_LabelTree}.
This may lead to improvements in performance and interpretability, but it comes with some caveats. In particular, it is not always available or straightforward to obtain these semantic relationships. In \cite{R1_Deng_HEX}, for instance, to perform classification on ImageNet, the authors had to resort to another database (WordNet \cite{wordnet}). 

\section{Coded ResNeXt}
\label{sec:Coded_ResNeXt}
    \subsection{The block}
    \label{subsec:One_block}
    \begin{figure*}[t]
       \centering 
       \includegraphics[trim=0mm 2mm 0mm 2mm, clip,width=1.0\linewidth]{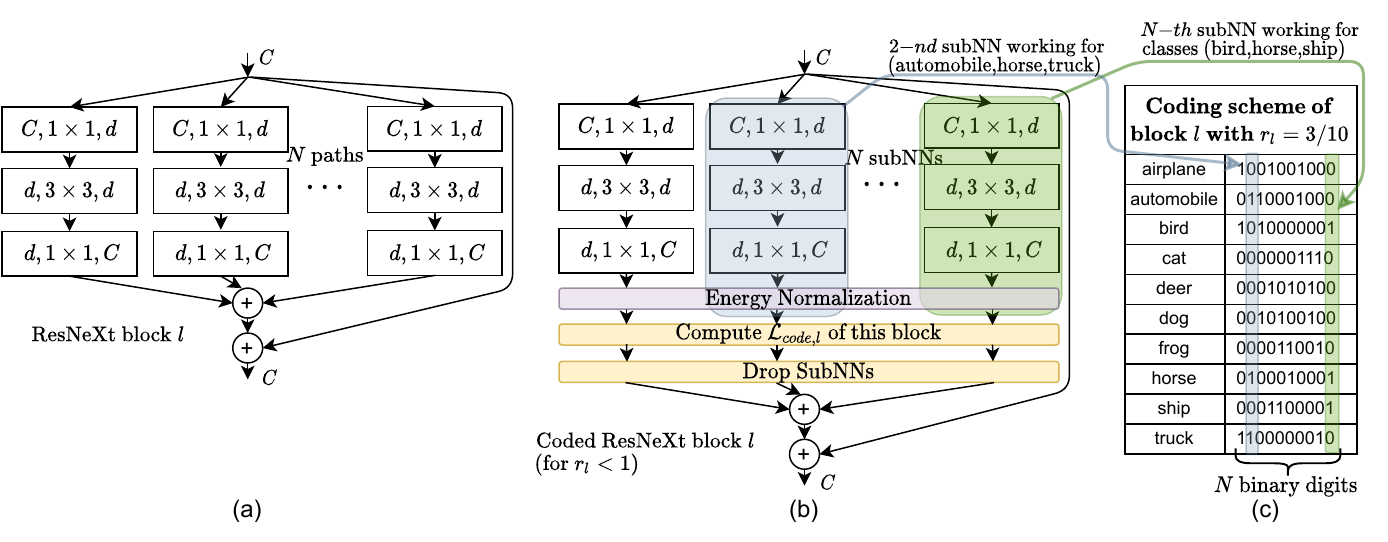}
       \caption{Building block of ResNeXt and the proposed variant. \textbf{(a)}: ResNeXt block. A layer is shown as (\# in channels, kernel size, \# out channels). \textbf{(b)}: Coded ResNeXt block. With light violet color we depict the architectural addition and with beige the algorithmic ones. The energy normalization keeps the total sum of the subNNs' output energies constant. Each subNN's output can be zeroed by the dropSubNNs operation with probability $p_{drop}$. Depending on the class of the sample, the loss $\mathcal{L}_{code,l}$ pushes the total energy to be allocated among subNNs with a specific order. \textbf{(c)}: An example of the order we used for CIFAR-10. We name this table as the coding scheme of the Coded ResNeXt block; this is defined before training and is kept fixed afterwards. To design the coding scheme we follow some general rules described in \cref{sec:How_to_code}. The ratio $r_l=3/10$ means that for the $l$-th block $N_{act,l}=3$ out of $N=10$ subNNs will work for each class. The $\mathcal{L}_{code,l}$ tries to match the output energy of the subNNs to their corresponding digit, depending on the binary codeword of the class.}
       \label{fig:arc}
    \end{figure*}
    
The typical ResNeXt block \cite{ResNeXt} is depicted in \cref{fig:arc}(a). It takes input $x\in\mathbb{R}^{C\times H\times W}$ ($C$ is the number of input channels and $H$, $W$ are the height and width of the input planes, respectively) and outputs $y$ of the same dimensions. It consists of $N$ paths/branches ($N$ is called cardinality in \cite{ResNeXt}). 
Each branch, which is called \emph{sub-neural network} (subNN) here, performs transformation $\mathcal{T}_n, \; n\in\{1,\cdots N\}$ which are all aggregated together with the input $x$, giving the block's output $y$:
\begin{equation}
y = x+\sum_{n=1}^{N}\mathcal{T}_n(x).
\end{equation}


\subsubsection{Energy Normalization}
For the Coded ResNeXt block depicted in \cref{fig:arc}(b), the sole architectural change we introduce is the \emph{Energy Normalization} applied before aggregating the transformed inputs $\mathcal{T}_n(x)\in\mathbb{R}^{C\times H\times W}$. For convenience, let $t_n\coloneqq\mathcal{T}_n(x)$. If $(t_n)_{c,h,w}\in \mathbb{R}$ is the element of $t_n$ in position $(c,h,w)$, then we define function $\mathcal{E}$ as:
\begin{equation}
\mathcal{E}(t_n) = \frac{1}{CHW}\sum_{c=1}^{C}\sum_{h=1}^H\sum_{w=1}^W\big( (t_n)_{c,h,w}\big)^2, \label{eq:energy_funtion}
\end{equation}
which gives the mean energy of the output signal of the $n$-th subNN. Energy Normalization simply divides the outputs of all branches by a scalar value equal to the square root of the total mean energy, i.e, 
\begin{equation}
\bar{t}_n = \frac{t_n}{\displaystyle\sqrt{\mathcal{E}_{avg}}} \textrm{ with } \mathcal{E}_{avg} = \frac{1}{N}\sum_{i=1}^N \mathcal{E}(t_i), \forall n\in \{1,\cdots,N\}.
\label{eq:Energy_Normalization}
\end{equation}
Given that $\mathcal{E}(ax)=a^2\mathcal{E}(x)$ for scalar $a\in \mathbb{R}_{\geq 0}$, it is easy to see that this step normalizes the total energy, since after it, the sum of the energy of all subNNs becomes $\sum_{n=1}^N \mathcal{E}(\bar{t_n})= N$. 
        
\subsubsection{Coding Loss}\label{sec:coding loss}
We present here our first algorithmic addition. After the Energy Normalization, we compute a novel loss function, coined \textit{coding loss} $\mathcal{L}_{code}$. Consider a classification problem of $K$ classes. Let $l$ be the index of the position of a ResNeXt block within the network. As seen in \cref{fig:arc}(c), for that block, we assign to each class a binary codeword $w_{l,k}, k\in\{1,\cdots, K\}$ of length $N$, indicating which subNNs we want to activate for that class. If the $n$-th subNN operates for class $k$, then the $n$-th digit of $w_{l,k}$ is $(w_{l,k})_n=1$, and $(w_{l,k})_n=0$ otherwise. To ensure that each class receives the same number $N_{act,l}$ of operating subNNs, all $K$ codewords are designed with exactly $N_{act,l}$ ones. 
We define the ratio
\begin{equation}
r_l = \frac{N_{act,l}}{N},
\end{equation}
which measures how much each class utilizes the block's total computational resources. We term the mapping of the classes to codewords, as in \cref{fig:arc}(c), the \textit{coding scheme} of the block. 
        
Given an input of class $k$, the coding loss forces the mean energies of the subNNs that are inactive for class $k$ to zero and those of the active subNNs to positive values. 
The coding loss for the $l$-th block is 
\begin{equation}
\mathcal{L}_{code,l} = \frac{1}{N}\sum_{n=1}^N(r_l\mathcal{E}(\bar{t_n})-(w_{l,k})_n)^4.\label{eq:Loss_code}
\end{equation}
Note that after the Energy Normalization, the total subNNs mean energy is $\sum_{n=1}^N \mathcal{E}(\bar{t_n})=N$, while the codeword has $N_{act,l}=r_lN$ ones, hence we multiply $\mathcal{E}(\bar{t_l})$ by $r_l$. 

We remark that the choice of setting the \textit{exponent to $4$} is carefully made. For example, setting it to $2$, the accuracy for CIFAR-10 drops from $94.4\%$ to $93.1\%$, which further drops to $87.1\%$ if the absolute value is used. An exponent of $2$ is much more demanding than the one of $4$ on matching precisely the output energies to the rules of the coding scheme, and therefore it seems to considerably restrict the flexibility of the function of the subNNs, degrading in turn the overall performance. This trend is exacerbated with using the absolute value. We observed as well the same behavior in CIFAR-100.
    
\subsubsection{DropSubNNs}
The second algorithmic addition is a type of dropout \cite{dropout}, similar to techniques such as SpatialDropout \cite{spatialDropout}, StochasticDepth \cite{stochastic_depth}, and DropPath \cite{Fractalnet_DropPath}. Seeing each subNN as one more complicated neuron, we apply dropout to it, so its output is zeroed with a fixed probability $p_{drop}$. This method is coined as \textit{DropSubNNs}. Our aim is to reduce the ``co-adaptation'' effect \cite{dropout} on the subNN level, according to which subNNs collaborate in groups instead of trying to independently produce useful features. In our implementation, we apply the same random mask to all blocks that have the same coding scheme. 


\subsection{The Network}
\label{subsec:The_network}
The complete network is constructed as a sequence of blocks. The Energy Normalization, $\mathcal{L}_{code,l}$, and dropSubNNs are applied only to blocks whose subNNs we want to specialize in some subsets of classes. Thus, for blocks with $r_l=N/N=1$, we use the conventional ResNeXt block as in \cref{fig:arc}(a). In that sense, the ResNeXt model is a Coded ResNeXt model where all blocks have $r_l=N/N$. 
    
\subsubsection{Coding Scheme Construction}\label{sec:How_to_code}
We remark that the coding scheme is constructed before training,  and that the subNNs are trained to comply with this fixed, predefined, scheme. 
In general, the coding scheme can be arbitrary, and can possibly incorporate semantic similarities between classes. 
However, 
we aim to make a proof of concept where \textit{it is possible to specialize subNNs to subsets of classes defined before training, even in the case when the classes within those subsets may not be semantically related}. Specifically, we found that even when the coding schemes are designed in an agnostic way with respect to the nature of the classes, good performance is guaranteed if some general construction rules are followed.

We construct one coding scheme per ratio $r_l$ so that a coding scheme is uniquely characterized by the ratio $r_l$ and any two blocks $l, l'$ with $r_l=r_{l'}$ have exactly the same coding scheme. A general rule we follow is that the deeper in the network a block is (i.e., the larger $l$ is), the smaller is the $r_l$ assigned. The first blocks have $r_l=N/N$ so that their subNNs produce low-level features, potentially useful for recognizing any of the classes. Deeper blocks have smaller $r_l$ so that their subNNs specialize on a subset of classes.\nolink{\footnote{In fact, the last linear layer of the ResNeXt can be seen as $K$ subNNs, each performing a simple linear combination, and the coding scheme has the lowest possible ratio $r_l=1/K$ (i.e., codewords are one-hot vectors).}} This rule not only is intuitive, but also works better in practice. For instance, in CIFAR-100, changing the proposed ratios ($20/20$, $8/20$, $4/20$) to ($8/20$, $8/20$, $8/20$) drops the accuracy from $78.8\%$ to $77.9\%$, and inverting the order into ($4/20$, $8/20$, $20/20$) gives $76.7\%$. 
                
Given a block $l$ with ratio $r_l$, we would like the coding scheme to satisfy the following three rules:
\begin{enumerate}[label=\Alph*.,leftmargin=*]
\setcounter{enumi}{0}
\item The number of ``1''s must be equal to $N_{act,l}=r_l N$ with $N$ being the codeword length.
\end{enumerate}
Moreover, we want to avoid under- or over-utilizing any subNN, in the sense of assigning too few or too many classes for it to process. As a result, the second rule is:
\begin{enumerate}[label=\Alph*.,leftmargin=*]
\setcounter{enumi}{1}
\item Seeing the coding scheme as a binary table, as in \cref{fig:arc}(c), the sum of each column should be approximately the same. 
\end{enumerate}
Finally, we aim at making the set of subNNs dedicated to work for a class, to be as different as possible from the sets assigned to the rest of the classes. This translates to:
\begin{enumerate}[label=\Alph*.,leftmargin=*]
\setcounter{enumi}{2}
\item The minimum Hamming distance 
 between all pairs of codewords should be as high as possible.
\end{enumerate}
Given $r_l$ and $N$, many coding schemes that follow the above rules may exist. For example, permuting the rows and/or the columns of the binary matrix in \cref{fig:arc}(c) gives new valid coding schemes. We experimentally checked (on CIFAR-10/100) that any scheme that satisfies the above properties provides similar results. 
Wanting to find $K$ binary codewords of length $N$ that only satisfy rule C is already an NP-Hard problem and in our case there are two additional rules. For that, we resort to an heuristic algorithm, presented in \cref{sec:CodingSchemeAlg}, which finds good coding schemes according to the above rules and was used to generate the codes of all our experiments. On a high level, the algorithm first constructs the set of all binary codewords of length $N$ with $N_{act}$ ones (rule A). Second, it extracts from it a subset containing only codewords whose mutual Hamming distance is always higher than a given threshold (rule C). Finally, it extracts multiple combinations of $K$ codewords from that subset and checks which one is a good coding scheme in terms of how well rule B is satisfied. Finally, we notice that the above coding scheme has some interesting connections with constant-weight codes.

\begin{table}
\renewcommand{\arraystretch}{1.4}
\centering
\footnotesize
\begin{tabular}{p{0.17cm}|c|c|c}
\hline
\parbox[t]{1.2mm}{\multirow{3}{*}{\rotatebox[origin=c]{90}{\hspace{28pt}\footnotesize{stage}}}}
&\thead{\footnotesize\textbf{Coded ResNeXt-29  (10$\times$11d)}\\ for CIFAR-10} 
& \thead{ \footnotesize\textbf{Coded ResNeXt-29 (20$\times$6d)}\\ for CIFAR-100 }
&\thead{ \footnotesize\textbf{Coded ResNeXt-50 (32$\times$4d)}\\ for ImageNet}\\
\hline\hline
s0  & conv $3{\times}3, 64$  & conv $3{\times}3, 64$ & \thead{\footnotesize conv $7{\times}7, 64$, str. 2, \\ $3{\times}3$ max pool, str. 2}\\
\hline
s1 & $[ 256, 11, 10/10 ]{\times }3$ & $[ 256, 6, 20/20]\times 3$ & $[256, 4, 32/32 ]{\times} 3$ \\
\hline
s2 & $[ 512, 22, \mathbf{5/10}  ]{\times }3$ & $[ 512, 12, \mathbf{8/20}]\times 3$ & $[512, 8, 32/32 ]{\times} 4$ \\
\hline
s3 & $[ 1024, 44, \mathbf{3/10}  ]{\times }3$ & $[ 1024, 24, \mathbf{4/20}]\times 3$ & $[1024, 16, \mathbf{16/32} ]{\times} 6$ \\
\hline
s4 & global avg. pool, 10-d fc & global avg. pool, 100-d fc  & $[2048, 32, \mathbf{8/32} ]{\times} 3$ \\
\cline{1-1}\cline{4-4}
&  & &  global avg. pool,\footnotesize{1000-d fc}\\
\hline
\end{tabular}
 \vspace{2pt}
\caption{Architecture for each dataset. A block is described by $[C_{out}, d, N_{act}/N]$, with $C_{out}$ being the number of channels it outputs and $d$ being the bottleneck width. For CIFAR architectures, stages s1, s2, s3 have approximately $0.2$, $0.9$, $3.5$ million parameters, respectively (in total $4.7M$). For ImageNet, s1, s2, s3 and s4 have  $0.2M$, $1.2M$, $7.0M$ and $14.5M$, respectively (in total $25.0M$).}
\label{tab:architectures}
\end{table}
       
\subsubsection{Architecture and Total Loss}
We succinctly describe a Coded ResNeXt block as $[C_{out}, d, r_l]$, with $C_{out}$ being the number of channels the block outputs and $d$ being the bottleneck width as in ResNeXt \cite{ResNeXt}. A conventional ResNeXt block is expressed as $[C_{out}, d, N/N]$. Following \cite{ResNeXt}, given the number of subNNs $N$, the bottleneck width $d$ is determined so that the blocks have about the same number of parameters and FLOPs as the corresponding blocks of the original ResNet bottleneck architecture \cite{ResNets}. 
\Cref{tab:architectures} presents the networks trained for CIFAR-10 (C10), CIFAR-100 (C100) \cite{CIFAR}, and ImageNet-1k (IN) \cite{Imagenet} classification datasets. In CIFAR-10/100 we chose $N$ to be small yet sufficiently high to enable reducing $r_l$ to less than $0.25$ and still obtaining a coding scheme with minimum Hamming distance not less than 4. For ImageNet we used the default values of ResNeXt-50. Remarkably, even though the number of classes increases exponentially across datasets ($K\in\{10,100,1000\}$), a strong coding scheme can be found to efficiently share the subNNs between classes, so that (a) random pairs of classes are assigned to very different subsets of subNNs; and (b) only a linear increase of the number of subNNs ($N\in\{10,20,32\}$) is needed.
            
Let $\mathcal{L}_{class}$ be the conventional cross entropy loss and $B_{code}$ be the set of indices $l$ pointing to the blocks with ratio $r_l<1$. 
Let $\mu$ be a loss-balancing constant; then the total loss used to train the network is
        \begin{equation}
            \mathcal{L}_{tot} = \mathcal{L}_{class} + \mu \sum_{l\in B_{code}} \mathcal{L}_{code,l}.\label{eq:total_loss}
        \end{equation}

\section{Experiments}
In this section, we present experimental results to assess the performance of the proposed Coded ResNeXt. First, we show that our algorithm achieves subNN specialization. To demonstrate this we show that when the subNNs specialized on the class of interest are removed, the performance degrades, whereas it remains the same or even improves when the subNNs removed are not specialized for that class. To further prove the specialization, given a class, we keep only the subNNs assigned to that class. That way, we retrieve a lightweight single-purpose binary classifier, accurately deciding whether the input sample belongs to the class or not. Finally, we show that it is possible to get good predictions from intermediate blocks without evaluating the whole network. Those predictions can also be used to provide confidence on whether the final network's prediction is correct. 
    
    \subsection{Setup and Validation Accuracy}\label{sec:Experiments_setup}
     \begin{wraptable}{r}{7.8cm}
        \vspace{-0.5cm}
      \centering
      \begin{tabular}{p{1.8cm}|c|c||c}
         & ($\mu,p_{drop}$) & \thead{ Coded\\ ResNeXt} & ResNeXt\\
        \toprule
        CIFAR-10 & (6, 0.1) &  \textbf{94.41\%}   & 93.66\% \\
        \hline
        CIFAR-100 & (6, 0.1) &  \textbf{78.76\%}  & 76.86\% \\
        \hline
        ImageNet & (2, 0.1) &  \textbf{80.24\%}  & 79.50\% \\
        \hline
      \end{tabular}
      \caption{Default hyperparameters and validation accuracy. For Coded ResNeXt on ImageNet $80.24\%$ is the mean of $3$ runs, which gave almost identical results $(80.21\%, 80.25\%, 80.26\%)$.}
      \label{tab:hyperparameters_accuracies}
        \vspace{-0.25cm}
    \end{wraptable}
        
    In order to make a fair comparison with ResNeXt, on ImageNet \cite{Imagenet} we follow the training process proposed by timm library \cite{timm_ResNeXt}. The epochs are $250$ (first $5$ as warmup \cite{lr_rules_of_thumb} and last $10$ cooling down), the batch size is $1536$, and the learning rate $0.6$. RandAugment \cite{Randaugment}  of $2$ layers and magnitude $7$ (varied with a standard deviation of $0.5$) is used and also random erasing augmentation \cite{Random_erase} with probability $0.4$ and $3$ recounts. 
    We diverge from the timm's proposed process only on the resolution of the input \emph{training} images. We reduce the resolution from 224 to 160, since on the TPU-v2 of Google Colab (the platform used for our experiments) the training would   take more than three weeks. 
    Still, timm reports $79.77\%$ accuracy for the ResNeXt-50, which is clearly smaller than the one of Coded ResNeXt-50, despite being trained with lower resolution. In \cref{sec:ImplementationDetails} we provide further details, including the training procedure for CIFAR.
    
    In \Cref{tab:hyperparameters_accuracies} we compare the accuracy of Coded ResNeXt against the corresponding ResNeXt (i.e., when setting all ratios $r_l$ equal to 1). We observe a clear improvement in accuracy across all datasets. Surprisingly, forcing the subNNs to specialize to specific set of classes yields significant gains even if the assignment of classes to subNNs is done in a way agnostic to the semantics of the classes.  
    \Cref{tab:hyperparameters_accuracies} presents the default values used for the introduced hyperparameters $(\mu,p_{drop})$ and the achieved validation accuracy. In \cref{sec:ablation} we perform an ablation study on those hyperparameters. 

    \subsection{Specialization}
     \begin{figure*}[t]
      \centering
      \begin{subfigure}{0.324\linewidth}
        \includegraphics[width=1.05\columnwidth]{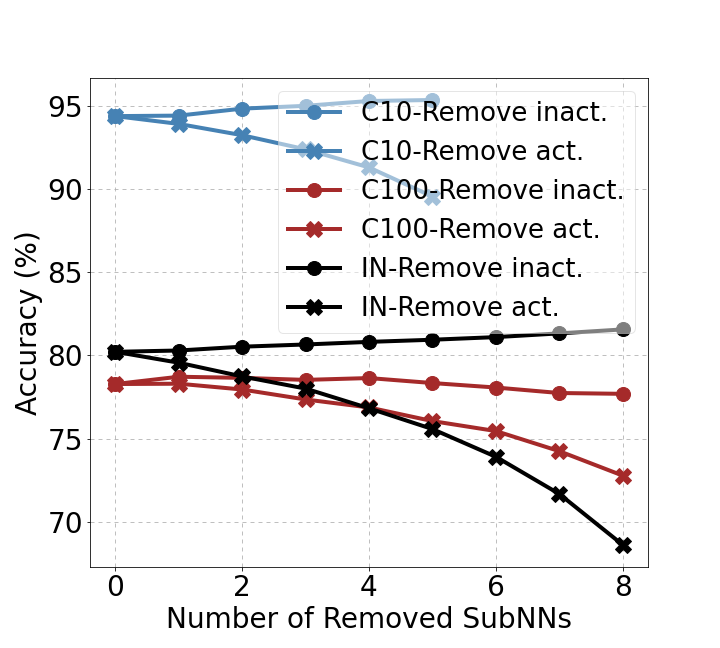}
       \caption{ Removing subNNs of a block.}
       \label{fig:RemovingSubNNs_GivenBlock}
      \end{subfigure}
      \hfill
      \begin{subfigure}{0.324\linewidth}
        \includegraphics[width=1.05\columnwidth]{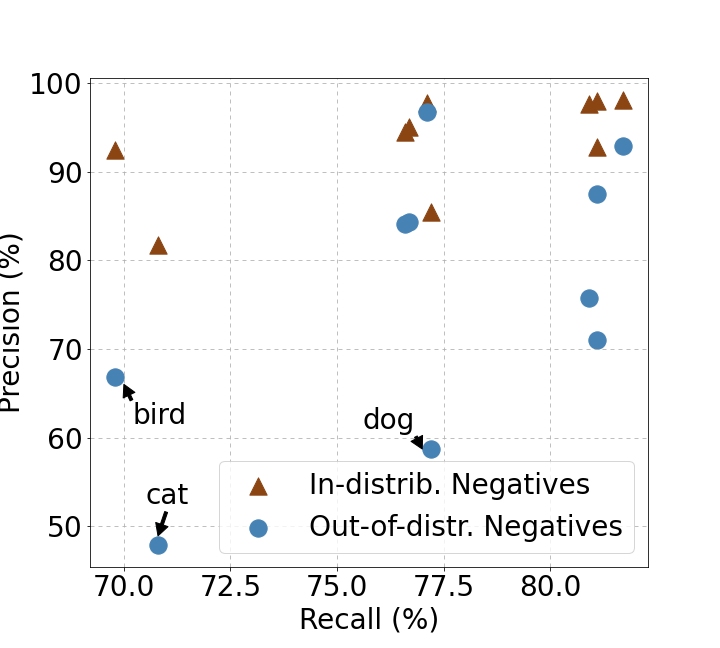}
       \caption{Precision-Recall on CIFAR-10.}
       \label{fig:PrecisionRecall_CIFAR10}
      \end{subfigure}
      \hfill
      \begin{subfigure}{0.324\linewidth}
        \includegraphics[width=1.05\columnwidth]{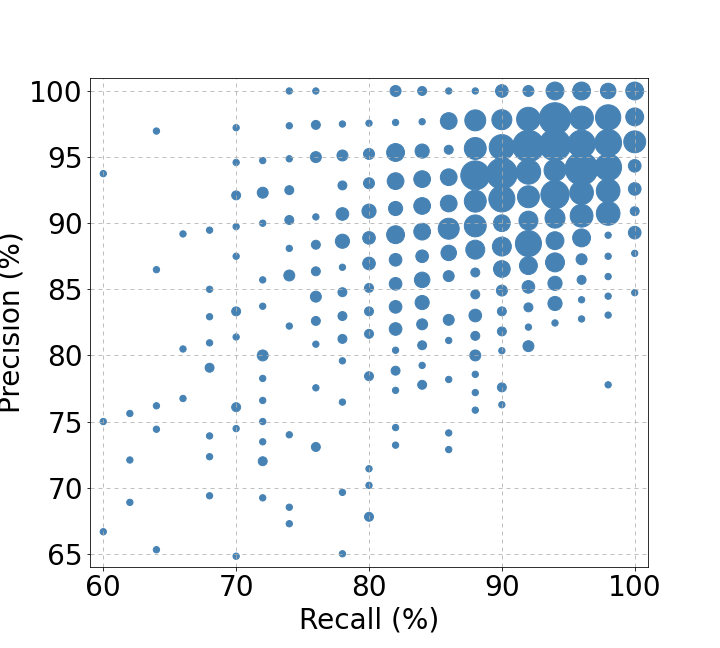}
       \caption{Precision-Recall on ImageNet.}
       \label{fig:PrecisionRecall_ImageNet}
      \end{subfigure}
      \caption{Demonstrating the specialization of subNNs to their assigned set of classes. \textbf{(a)} Performance when removing active versus inactive subNNs from a specific block. \textbf{(b)} Precision-Recall from all extracted binary classifiers trained on CIFAR-10. Out-of-distribution negatives are the validation set of CIFAR-100. \textbf{(c)} Precision-Recall from all extracted binary classifiers trained on ImageNet. The larger the marker, the more points fall into that area.}
      \label{fig:short}
    \end{figure*}
        
    A key idea of our work is to specialize each subNN to specific subset of classes; hence the first experiment is designed to test whether our architecture succeeds in achieving specialization. Assuming a subNN is assigned to activate for some class, if this subNN helps indeed on the classification process of images belonging to that class, removing this active subNN should negatively impact this process. On the other hand, if that subNN is not assigned to that class, then it should remain inactive during the process, so removing it should have no impact (degradation) on the performance. 
        
    For the first experiment we pick a block $l$ from which we randomly remove subNNs\footnote{Removing a subNN from a block in this architecture is equivalent to zeroing all of its parameters or to zeroing its output before the Energy Normalization.} in two ways. Given the class of the input image sampled from the validation set, the first way randomly removes $k{\leq} N_{act,l}$ subNNs from the set of active for that class subNNs. The second way randomly removes $k{\leq} N-N_{act,l}$ subNNs from the (complementary) set of inactive subNNs for that class. 
    For illustration, in \cref{fig:RemovingSubNNs_GivenBlock} we pick the last block of stage s2 in the architecture for CIFAR (see \cref{tab:architectures}) and the second of stage s4 for that for ImageNet. Figures with respect to other blocks are presented in \cref{sec:BinaryClassifiersAdditionalPlots}.
    
    In \cref{fig:RemovingSubNNs_GivenBlock}, we observe the same behavior across all datasets, which confirms that the more active subNNs are removed, the more the performance degrades. Interestingly, when removing inactive ones, the accuracy tends to increase. Our interpretation is that even though the inactive subNNs are trained to output zero signal, this is never perfectly achieved in practice and their output always interferes with that of the active subNNs. Thus, taking out the interferers could improve accuracy. Note that this higher accuracy of the neural network is not actually achievable since to remove a subNN we need to know a priori the class of the input so as to know the set of (in)active subNNs for that class. Finally, we remark that even if all active subNNs are removed from one block, the performance does not necessarily plummet. We believe that the reason behind this is that, in that case, information can still pass from the previous block to the next one through the skip connection.

    \subsection{Binary Classifier}
    Having confirmed that the subNNs specialize on their assigned subset of classes, we proceed with testing this property to the extreme. For that, instead of randomly removing few subNNs from one block, given a class $k$, we remove from \emph{all} blocks all subNNs not assigned to class $k$. The rationale behind this is to check whether by keeping only the subNNs specialized on one class we can obtain a binary classifier capable of recognizing that class among the others. 
        
    \begin{figure}[t]
      \centering 
      \begin{subfigure}{0.32\linewidth}
        \includegraphics[trim=20mm 0mm 20mm 20mm, clip,width=1.0\columnwidth]{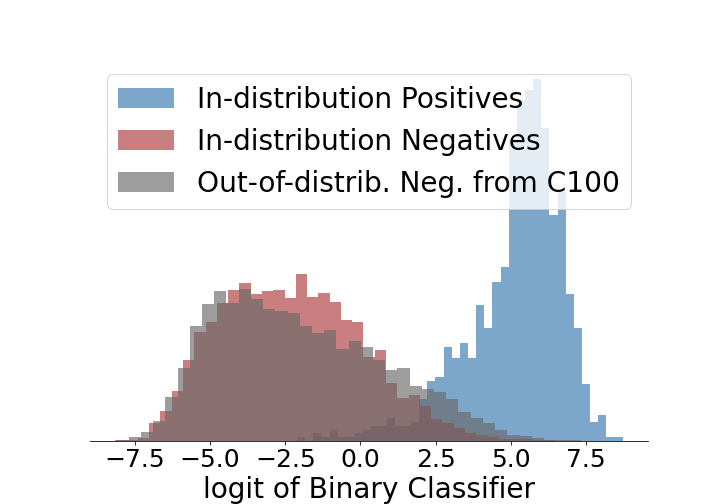}
        \label{fig:BinaryClassifier_cifar10}
      \end{subfigure}
      \hfill
      \begin{subfigure}{0.32\linewidth}  
        \includegraphics[trim=20mm 0mm 20mm 30mm, clip,width=1.0\columnwidth]{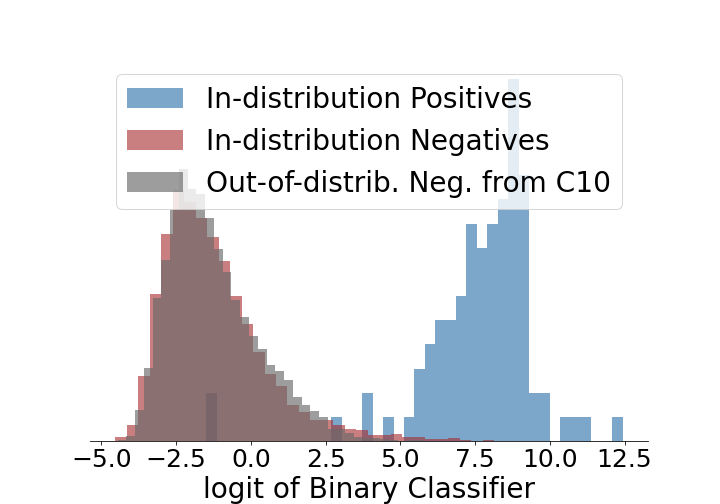}
        \label{fig:BinaryClassifier_cifar100}
      \end{subfigure}
      \hfill
      \begin{subfigure}{0.32\linewidth}
        \includegraphics[trim=20mm 0mm 20mm 20mm, clip,width=1.0\columnwidth]{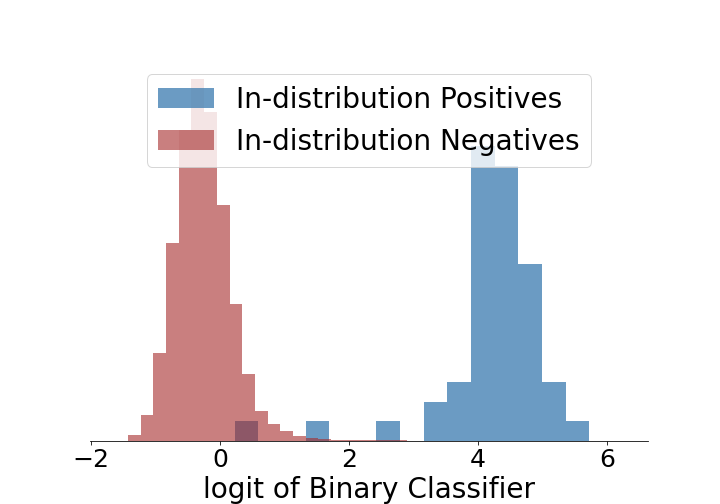}
        \label{fig:BinaryClassifier_ImageNet}
      \end{subfigure}
      \caption{Output distribution of Binary Classifier (BC) of the first class (airplanes, apples, tench) of each dataset (CIFAR-10, CIFAR-100, ImageNet, respectively).}
      \label{fig:BinaryClassifiers}
    \end{figure}

    In \cref{fig:BinaryClassifiers} we pick the first class of CIFAR-10/100 and ImageNet (``airplane'', ``apple'', and ``tench'', respectively) and remove all inactive subNNs for that class. We also remove from the final linear layer everything except its first row of parameters so as to keep only the first logit corresponding to the first class. That way, we retrieve a sub-model whose output is one-dimensional. \Cref{fig:BinaryClassifiers} depicts with blue the output distribution when inputting samples of the validation set belonging to the first class of the dataset (i.e., in-distribution positives), and with red when the samples belong to some other class (i.e., in-distribution negatives). Clearly, the extracted sub-models do operate as binary classifiers giving high output when fed with samples of the class for which they are specialized. To further showcase the specialization, for the sub-models trained on CIFAR-10 (resp. CIFAR-100) we input samples that belong to the validation set of CIFAR-100 (resp. CIFAR-10). Those are considered out-of-distribution (OOD) predictions, since the sub-model has never been trained on such samples. Nevertheless, as \cref{fig:BinaryClassifiers} shows, the extracted BC still perform  very well. A possible justification for the the good OOD performance of the extracted BC is the functional lottery ticket hypothesis \cite{R4_SubNetworks_OOD}, which states that every full network contains a subnetwork that can achieve better OOD performance.

    Therefore, \emph{with a single training of a large multi-purpose neural network, we can straightforwardly extract multiple single-purpose models that are considerably lighter ($38\%$, $27\%$, and $35\%$ of the initial parameters for CIFAR-10, CIFAR-100, and ImageNet architectures, respectively)}. Given a threshold distinguishing between positive and negative predictions, each of those models becomes a BC. We set that threshold to the value maximizing the F1-score of the BC when fed with samples from the training dataset. In \cref{fig:PrecisionRecall_CIFAR10,fig:PrecisionRecall_ImageNet}\footnote{The validation set of ImageNet has $50$ positives and $999{*}50{=}49950$ negatives per class. In \Cref{fig:PrecisionRecall_ImageNet} we consider only $9{*}50{=}450$ randomly selected negatives to compute the precision and recall. We do that (i) in order to keep the same ratio of positives versus negatives as in CIFAR-10 and allow comparison, and (ii) because the dataset is very skewed; e.g., even a very conservative threshold that misclassifies only $1\%$ of the negatives results into approximately $500$ false positives. Since they are only $50$ positives, the precision becomes $10\%$.}
    the  performance of the BCs (on the validation set and the out-of-distribution set) is depicted in precision-recall plots. Notably, for CIFAR-10, the worst performance is obtained by the BC for ``cats''  when fed with CIFAR-100's out-of-distribution samples. This seems reasonable, since we request from the classifier to distinguish cats from classes like leopard, lion, and tiger, but without having ``seen'' any sample of them during training.

    In ResNet, complete blocks can be removed without severely degrading the accuracy \cite{RemovingBlocks_ResNets}. Hence, it is reasonable to ask whether the conventional ResNeXt is also robust to the removal of subNNs and thus, good BCs can be extracted from it without the need for our proposed modifications. Interestingly, this is not the case and the answer is negative. In CIFAR-10 for instance, the extracted BCs from the Coded ResNeXt give on average precision $93\%$ and recall $77\%$ (F1-score $F_1=84\%$). Attempting to extract likewise BCs from a ResNeXt leads to precision $13\%$ and recall $56\%$. Finally, given a class, the complete ResNeXt architecture can be seen as a BC by considering its output to be only the corresponding logit. Comparing such BCs to the extracted BCs seems unfair since the extracted BCs not only have $2.5$ times fewer parameters, but also have never been trained as independent models. Nonetheless, this may serve as a baseline. For CIFAR-10 this baseline gives BCs with average precision $79\%$ and recall $94\%$  ($F_1=86\%$). Additional details and plots are provided in \cref{sec:BinaryClassifiersAdditionalPlots}. 

    \subsubsection{Why ResNeXt?}\label{sec:whyResNeXt}
    In this subsection, we provide insights on why the subNNs achieve specialization and we highlight why ResNeXt serves as the appropriate architecture upon which to build our idea. The objective of our work is to construct networks in which the per-class information is forced to flow through specific paths (determined here by the coding schemes). To achieve this, we employ (i) an operation (energy normalization) that limits how many subNNs can be activated; and (ii) a loss function forcing which ones should be activated. Intuitively, those operations should suffice for constraining the information to flow through the active subNNs. A natural question that arises is how accurate this is. 
    Let us assume that it is accurate. Then, keeping those operations unaltered and changing only the way the subNNs' outputs are passed to the subsequent blocks should not impact the flow of information. However, if instead of aggregating them by \textit{summation}, they are \textit{concatenated}, the performance of the extracted BCs becomes poor (precision $< 20\%$). It seems that the concatenation inhibits the ``information'' to pass only through the designated paths, since the performance degrades when inactive subNNs (which in theory should not participate in those paths) are removed. Let us see why.

    When concatenating the outputs of the $l$-th block, the information about which are the inactive subNNs is preserved, thus the $(l{+}1)$-th block may depend its operation on which subNNs of the $l$-th block provide zero output. This allows information to ``leak'' from the inactive subNNs. On the contrary, the information that some subNNs provide zero output is lost when adding them to the final output of the block. For that reason, the ResNeXt architecture (which aggregates the outputs by summation) is very well suited for developing our idea of controlling the information paths. Interestingly, another popular block called MBConv proposed for MobileNet-V2 \cite{R3_MobilenetV2} (and later used for EfficientNets \cite{R3_efficientnet}) bears a resemblance to ResNeXt block and is also a good candidate for incorporating our ideas. We elaborate more in \cref{sec:limitations}.

    
    
    \begin{figure}[h]
      \begin{minipage}[b]{0.45\textwidth}
          \centering
           \includegraphics[width=1.0\columnwidth]{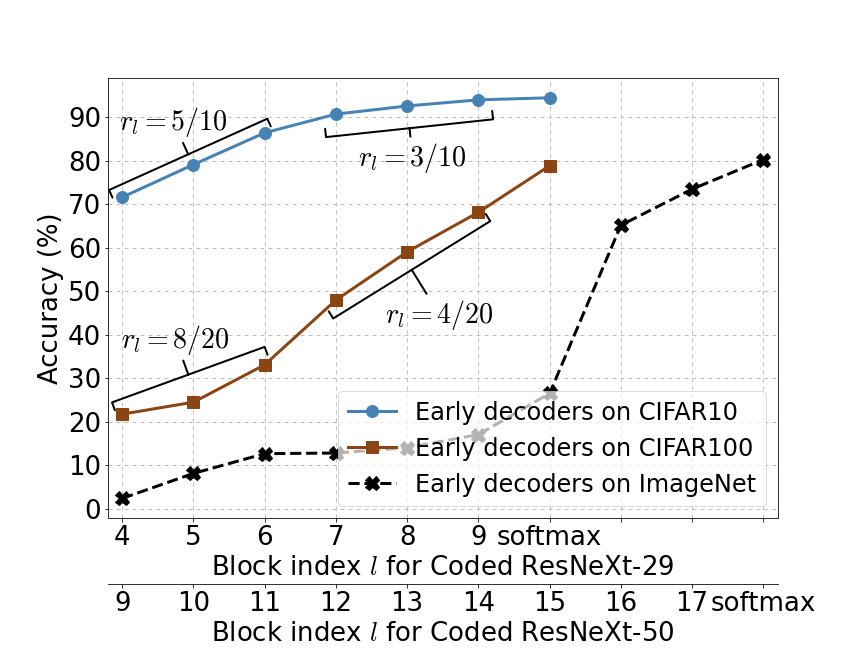}
           \captionof{figure}{Accuracy of the early decoders.}
           \label{fig:EarlyDecoders}
      \end{minipage}
      \hfill
      \begin{minipage}[b]{0.52\textwidth}  
            \hspace{-0.46cm}
            \renewcommand{\arraystretch}{1.2}
            \small
            \begin{tabular}{c|c|c|c|c|}
            \multicolumn{2}{c|}{} & \rotatebox[origin=c]{0}{\small CIFAR-10} & \rotatebox[origin=c]{0}{\small CIFAR-100} & \rotatebox[origin=c]{0}{\small ImageNet} \\
            \cline{2-5}%
            \parbox[t]{9mm}{\multirow{7}{*}{\rotatebox[origin=c]{90}{\thead{ number of early \\decoders agreeing\\ with final prediction}}}}
            & 0         & 52.0\% & 48.7\% & 46.6\% \\
            \cline{2-5}
            & 1         & 67.7\% & 69.5\% & 70.5\% \\
            \cline{2-5}
            & 2         & 77.2\% & 77.2\% & 87.3\% \\
            \cline{2-5}
            & 3         & 83.2\% & 87.7\% & 90.7\% \\
            \cline{2-5}
            & 4         & 89.9\% & 90.7\% & 90.9\% \\
            \cline{2-5}
            & 5         & 92.1\% & 90.6\% & 91.1\%\\
            \cline{2-5}
            & $\geq 6$  & 98.1\% & 95.8\% & 92.6\%\\
            \cline{2-5}
          \end{tabular}
          \captionof{table}{Accuracy of final prediction ($\%$) given the number of early decoders giving the same prediction.}
        \label{tab:conditional_accuracies}
    \end{minipage}
    \end{figure}
    
    \subsection{Early Decoding}\label{sec:EarlyDecoder}
    Coded ResNeXt improves accuracy over ResNeXt while enabling the extraction of multiple lighter single-purpose models with a single training and providing transparency on how information flows throughout the network. Here we show that leveraging coding theory to design when and which subNNs should be activated allows exploiting Coded ResNeXt in other ways.

    Given block $l$ with $r_l<1$, i.e., $l\in B_{code}$, the coding scheme maps each class $k\in\{1,\cdots,K\}$ in \textit{one-to-one} fashion to a codeword $w_{l,k}$ and then the training pushes the energies $v_l\in \mathbb{R}_{\geq 0}^N$ of the block's subNNs output to match that codeword. This allows for each $l\in B_{code}$ to measure the vector $v_l\in \mathbb{R}_{\geq 0}^N$, find the codeword $w_{l,k}, k\in\{1,\cdots,K\}$ having the minimum distance to $v_l$, and consequently predict the class of the sample. As a result, each block $l\in B_{code}$ becomes an early decoder predicting label $\underset{k}{\arg \min} \; ||v_l-w_{l,k}||_2, \; k\in\{1,\cdots,K\}$, 
    with $||\cdot||_2$ being the L2 norm. 
    In \cref{fig:EarlyDecoders} we depict the accuracy of every block $l\in B_{code}$ when functioning as an early decoder. Interestingly, as a sample passes from one block to the next one, the probability of being correctly decoded increases. In \cref{sec:ablation} we show that as the coefficient $\mu$ of $\mathcal{L}_{code,l}$ increases, the early decoders get improved, but past a certain point this comes at the expense of the overall accuracy.
      
    
    Finally, we illustrate another possible utility of the early decoders. In \cref{tab:conditional_accuracies} we measure the accuracy of the network's final prediction given how many early decoders also provide the same prediction. It is obvious that the more decoders agree with the final prediction, the higher the probability to be correct. Therefore, early decoders provide a confidence estimation on the correctness of the network's prediction. Specifically, they can be a source of extra features used to improve state-of-the-art confidence calibration methods \cite{ConfidenceCalibration}. 
    We further analyze this in \cref{sec:confidence calibration}.

\section{Conclusion}
In this paper, we proposed a network architecture in which the information related to each class flows through distinctive and clearly defined paths. We depart from the ResNeXt architecture and apply few yet crucial modifications that allow achieving higher accuracy and have several additional attractive properties. First, we specialize each part of the model on a specific and fixed subset of classes, which --given a certain class-- enables to obtain a binary classifier for that class by keeping only the relevant parts of the network. Second, it allows to obtain early predictions without the need for entirely evaluating the network. Third, if fully evaluated, a confidence level can be produced on the correctness of the final prediction. In this work, we achieved specialization without having to rely on the semantic similarities between classes. Nonetheless, we conjecture that further gains (higher accuracy, lighter binary classifiers, smaller network, etc.) can be obtained by exploiting such similarities.
Finally, while in this work we have shown that our method outperforms ResNeXt in image classification, future directions may include comparing in other computer vision tasks such as detection, few-shot learning and robustness against adversarial attacks. 

\section*{Acknowledgements}
The authors are grateful to Angelos Katharopoulos and Marina Costantini for helpful discussions and remarks, and also thank the anonymous referees for their useful and constructive comments. The work of A. Avranas has been supported by the EURECOM-Huawei Chair on Advanced  Wireless Networks. M. Kountouris has received funding from the European Research Council (ERC) under the European Union’s Horizon 2020 research and innovation programme (Grant agreement No. 101003431).

{\small
\bibliographystyle{ieee_fullname}
\bibliography{egbib}
}

\section*{Checklist}
    \begin{enumerate}
    
    \item For all authors...
    \begin{enumerate}
      \item Do the main claims made in the abstract and introduction accurately reflect the paper's contributions and scope?
        \answerYes{}
      \item Did you describe the limitations of your work?
        \answerYes{}
      \item Did you discuss any potential negative societal impacts of your work?
        \answerNA{}
      \item Have you read the ethics review guidelines and ensured that your paper conforms to them?
        \answerYes{}
    \end{enumerate}

    \item If you are including theoretical results...
    \begin{enumerate}
      \item Did you state the full set of assumptions of all theoretical results?
        \answerNA{}
            \item Did you include complete proofs of all theoretical results?
        \answerNA{}
    \end{enumerate}

    \item If you ran experiments...
    \begin{enumerate}
      \item Did you include the code, data, and instructions needed to reproduce the main experimental results (either in the supplemental material or as a URL)?
        \answerYes{The datasets are publicly available and the code is included in the supplementary material.}
      \item Did you specify all the training details (e.g., data splits, hyperparameters, how they were chosen)?
        \answerYes{}
            \item Did you report error bars (e.g., with respect to the random seed after running experiments multiple times)?
        \answerYes{}
            \item Did you include the total amount of compute and the type of resources used (e.g., type of GPUs, internal cluster, or cloud provider)?
        \answerYes{}
    \end{enumerate}

    \item If you are using existing assets (e.g., code, data, models) or curating/releasing new assets...
    \begin{enumerate}
      \item If your work uses existing assets, did you cite the creators?
        \answerYes{}
      \item Did you mention the license of the assets?
        \answerYes{}
      \item Did you include any new assets either in the supplemental material or as a URL?
        \answerYes{}
      \item Did you discuss whether and how consent was obtained from people whose data you're using/curating?
        \answerNA{}
      \item Did you discuss whether the data you are using/curating contains personally identifiable information or offensive content?
        \answerNA{}
    \end{enumerate}

    \item If you used crowdsourcing or conducted research with human subjects...
    \begin{enumerate}
      \item Did you include the full text of instructions given to participants and screenshots, if applicable?
        \answerNA{}
      \item Did you describe any potential participant risks, with links to Institutional Review Board (IRB) approvals, if applicable?
        \answerNA{}
      \item Did you include the estimated hourly wage paid to participants and the total amount spent on participant compensation?
        \answerNA{}
    \end{enumerate}

    \end{enumerate}

\newpage

\appendix
\section{Coding schemes}
\label{sec:CodingSchemeAlg}

    In this section we present the construction methodology of coding schemes. 
    We drop the subscript $l$ from $r_l$ and $N_{act,l}$
    and show the general rule for an arbitrary block for which those values are given. 
    For the reader's convenience, we repeat the three rules the coding scheme should comply with. Given some block with ratio $r$ and number of branches/subNNs $N$, the rules are as follows:
    \begin{enumerate}[label=\Alph*.]
        \item The number of ``1''s must be equal to $N_{act}=r N$ with $N$ being the codeword length.
        \item Seeing the coding scheme as a binary table, with each row representing a class and each column a subNN as in \cref{tab:CIFAR10-codingSchemes}, we require the sum of each column to be approximately the same.\footnote{Note that due to the first rule the sum of each row is equal to $N_{act}$.} \label{rule:same_cardinality}
        \item The minimum Hamming distance across the pairs of codewords should be as high as possible.
    \end{enumerate}
    The first rule is \textit{mandatory} and we only consider codewords with number of ``1''s equal to $N_{act}$. 
    The other two rules would serve as \textit{guidelines} and we try to follow them to the maximum extent possible.
    Let $S_{min}$ (resp. $S_{max}$) be the sum of the columns with the minimum (resp. maximum) sum.
    The second rule is fully satisfied if $S_{min}=S_{max}$. 
    This is not realizable for all ratios $r$. The ratio $r$ must be chosen taking into account the number of classes $K$ as follows.
    The number of ``1''s in the binary table is $KN_{act}$. Assuming a coding scheme with $S_{min}=S_{max}=S_{opt}$, the number of ``1''s is also equal to $NS_{opt}$. 
    This brings the equality $KN_{act}=NS_{opt}\Leftrightarrow S_{opt}=rK\in\mathbb{N}$. 
    Hence, a necessary condition to be able to find a coding scheme with $S_{min}=S_{max}$ is that $r K\in\mathbb{N}$.

    The number of possible combinations for choosing $p$ elements from a set of $n$ distinct elements is given by $C(n,p)=\frac{n!}{p!(n-p)!}$. A coding scheme where each class is mapped to a distinct codeword exists if $C(N,rN)\geq K$. Therefore the chosen $r$ should satisfy this inequality.

    \begin{wraptable}{r}{7cm}
      \centering
      \begin{tabular}{c||c|c|}
         &  $r=5/10$ & $r=3/10$ \\
        \toprule
        airplane    & 1010100011 & 1001001000\\
        \hline
        automobile & 0101010101 & 0110001000 \\
        \hline
        bird      & 1101100010 & 1010000001 \\
        \hline
        cat      & 0011001101 &  0000001110 \\
        \hline
        deer     & 1010010101 & 0001010100   \\
        \hline
        dog      & 1001001110 &  0010100100 \\
        \hline
        frog     & 1011101000 &  0000110010 \\
        \hline
        horse    & 0100011110 & 0100010001 \\
        \hline
        ship    & 0110111000  &  0001100001 \\
        \hline
        truck   & 0100110011  & 1100000010 \\
        \hline
      \end{tabular}
      \caption{The coding schemes used in CIFAR-10.}
      \label{tab:CIFAR10-codingSchemes}
    \end{wraptable}
    
    Let $H_{min}$ be the minimum Hamming distance within the set of all possible pairs of codewords in the coding scheme. In CIFAR-10 for instance, as it can be verified from \cref{tab:CIFAR10-codingSchemes}, we have $H_{min}=4$ for both $r=5/10$ and $r=3/10$ (check the pair horse-ship). Obviously, the higher $C(N,rN)$ is, the larger the set of acceptable codewords to choose for the coding scheme is, and the larger the $H_{min}$ that can be achieved. Nevertheless, not every value of $H_{min}$ is achievable. There is a value above which there is no such coding scheme with $K$ codewords.
    
    After choosing $r, N$ such that $r K\in\mathbb{N}$ and $C(N,rN)\geq K$, we proceed to find the coding scheme. Finding a coding scheme that satisfies the three rules mentioned above is very challenging. Actually, there is no known way to compute even the ``basic'' function $A_2(N,\mathsf{d})$ that gives the maximum number of binary codewords of length $N$ with minimum Hamming distance $\mathsf{d}$. 
    Moreover, computing the function $\mathsf{D}(N,K)$ that gives the minimum possible Hamming distance of a coding scheme of $K$ codewords is even harder. 
    Using $\mathsf{D}(\cdot)$ one can evaluate $A_2(\cdot)$ through a binary search over $K$. In our case, the additional constraint of having $N_{act}$ ``1''s increases the difficulty. 
    Lastly, in addition to knowing the existence of such a coding scheme, we are interested in realizing it, i.e., generating a valid set of codewords for that scheme. 
    For that, we have to resort to heuristics 
    that satisfy to the largest extent the three aforementioned rules. 
    In \cref{alg:coding_scheme} we give the pseudocode of the algorithm used to generate the coding schemes of CIFAR-100 and ImageNet. 
    For CIFAR-10 the length of the codewords is $N=10$, which is small enough to allow for the use of a brute force approach similar to exhaustive search.
    
    \begin{algorithm}
        \caption{Algorithm for generating a coding scheme}\label{alg:coding_scheme}
        \begin{algorithmic}[1]
        \Require $K, N, N_{act}, H_{min}$
        \Function{MinHamDist}{codeword $w$, set $G$}
            \State $d\gets \infty$
            \For{$w_g$ in $G$}
                \If{HammingDistance$(w,w_g)<d$}
                    \State $d\gets \mathrm{HammingDistance}(w,w_g) $
                \EndIf
            \EndFor
            \State \Return $d$
        \EndFunction
        \item[]
        \Function{Score}{coding scheme $C$}
            \State $S_{min}\gets \min \{ \text{sum  column of coding scheme } C \}$
            \State $S_{max}\gets \max \{ \text{sum  column of coding scheme } C \}$
            \State \Return $S_{max}-S_{min}$ \Comment{The lower, the better}
        \EndFunction
        \item[]
        \State $L\gets$List of all codewords with $N_{act}$ ones  \label{line:set L}
        \State $L_{sorted}\gets \mathrm{sort}( L )$ \label{line:sort}
        \State $G\gets \{ \}$
        \For {$w$ in $L_{sorted}$}
            \If{\Call{MinHamDist}{$w, G$}$\geq H_{min}$}
                \State $G\gets G\cup \{w\}$
            \EndIf
        \EndFor
        \If{cardinality of $G<K$}
            \State exit \Comment{Unable to find a coding scheme}
        \EndIf
        \State $BestScore\gets \infty$
        \ForAll{$C \subseteq G$ with $|C|=K$}
            \State $score\gets$\Call{Score}{$C$}
            \If {$score = 0$}
                \State\Return{$C$} \Comment{Found solution satisfying rule C}
            \ElsIf{$score<BestScore$}
                \State $BestScore\gets score$
                \State $BestSchemeFound \gets C$
            \EndIf
        \EndFor
        \State\Return $BestSchemeFound$
        \end{algorithmic}
    \end{algorithm}
    
    \cref{alg:coding_scheme} starts by creating a list of length $C(N,rN)$ with all possible codewords satisfying rule A. 
    Each codeword can be mapped to the integer whose binary representation matches the codeword. 
    These integers are used to sort the list of codewords in line \ref{line:sort}.  
    This step is crucial, since randomly picking codewords is very inefficient for creating the set $G$ in the subsequent lines. 
    The set $G$ is a set of codewords in which all possible pairs of codewords belonging in this set have Hamming distance between them at least $H_{min}$. 
    The larger $G$ is, the easier it is to find a subset of cardinality $K$ that satisfies all three rules. 
    
    We now give some intuition on why picking codewords randomly from the set $L$ (see line \ref{line:set L}) would result in a much smaller set $G$ than the method proposed that picks them sequentially from $L_{sorted}$. 
    Consider the following analogy: imagine having disks with radius of $H_{min}$ instead of codewords. 
    The problem is to fit as many non-overlapping disks as possible inside a square. 
    If we start filling the square by randomly placing the disks inside the square, this will quickly result in no extra disk actually fitting within the space left by the already placed ones, even though there is still a lot of space unoccupied. 
    On the other hand, if the disks are placed in an ordered way, for example starting from the edges and progressively placing them as close as possible to the already placed disks, then many more disks will eventually fit. 
    
    \cref{alg:coding_scheme} is a simplified version of our implementation. In line 20, the number of possible sets $C$ that can be chosen from $G$ can be extremely large. 
    In that case, we resort to additional heuristics for picking only good candidates for $C$. Further details can be found in our Python code. 
    Finally, the coding scheme for CIFAR-100 with $r=8/20$ is retrieved using the arguments $(K, N, N_{act}, H_{min})=(100, 20, 8, 8)$ in the \cref{alg:coding_scheme} and with $r=4/20$ using  $(K, N, N_{act}, H_{min})=(100, 20, 4, 4)$. 
    For both ratios the coding schemes found entirely satisfy rule \ref{rule:same_cardinality}, i.e., $S_{min}=S_{max}$. The coding scheme for ImageNet with $r=16/32$ is retrieved using arguments $(K, N, N_{act}, H_{min})=(1000, 32, 16, 10)$ achieving $S_{min}=499\approx S_{max}=501$. For the coding scheme with $r=8/32$, we use arguments $(K, N, N_{act}, H_{min})=(1000, 32, 8, 6)$, achieving $S_{min}=249\approx S_{max}=251$.

\section{Implementation details and computational cost}
\label{sec:ImplementationDetails}
    We present here some additional implementation details and show the computational cost for training a Coded ResNeXt network with respect to the cost of training a conventional ResNeXt. 
    
    As shown in \cref{fig:arc}(a) and (b), each path/subNN of the ResNeXt and Coded ResNeXt blocks consists of three layers. Each of the first two layers is composed of a convolutional operation followed by a batch normalization (BN) \cite{BatchNorm}, and a rectified linear unit (ReLU) \cite{Relu}. For the ResNeXt block, in the last layer, after the convolutional operation the output of all paths/subNNs is aggregated by summation, followed by BN and ReLU.
    Similarly, for Coded ResNeXt, there is a BN and a ReLU operation that come after aggregating the output of all subNNs.
    In \cref{fig:arc}(a) and (b) those two operations would be depicted between the two summations.

    We run all our experiments on Google's Colab TPU-v2 ($N_w = 8$ cores with precision bfloat16). We used PyTorch's implementation of stochastic gradient descent with Nesterov momentum \cite{sgd_nesterov,sgd_adaptation_sutskever} equal to $0.9$. We used the cosine scheduler \cite{cosine_scheduler} that decayed the learning rate until $10^{-5}$. 
    
    For CIFAR datasets the batch size is picked relatively high to harness TPU speed; the batch size per core is set to $B_w=64$ (i.e., effective $512$). Training on CIFAR is performed for $300$ epochs with initial learning rate $0.1$ and weight decay \cite{WeightDecay1992} equal to $5\cdot 10^{-4}$. 
    For data augmentation we used RandAugment \cite{Randaugment} with $(N_{aug},M_{aug})=(3,4)$ for CIFAR-10 and $(1,2)$ for CIFAR-100\footnote{Those values are chosen in \cite{Randaugment} for Wide-ResNet-28-2 \cite{WideResNets} which, out of all models presented in that work, seems the most similar to ResNeXt-29.}, which we applied after the standard pad-and-crop and horizontal flips with probability $0.5$.
    
    As mentioned in \cref{sec:Experiments_setup}, in order to make a fair comparison with ResNeXt, on ImageNet \cite{Imagenet} we follow the training process proposed by the timm library \cite{timm_ResNeXt}\footnote{It is possible that using more recent guidelines like the ones proposed for ResNet in \cite{Revisiting_ResNets_guidelines} could give even higher accuracy. However, in this case, we would not have available an already publicly reported accuracy for ResNeXt. Therefore, we prefer to follow the recipe of timm's library.}. 
    The epochs are $250$ (first $5$ as warmup \cite{lr_rules_of_thumb} and last $10$ cooling down), the effective batch size is $1536$ ($192$ per core), the learning rate is $0.6$, and weight decay is  $10^{-4}$. 
    The input image is first randomly resized and cropped using the standard values of scale and ratio \cite{InceptionV1} and then horizontally flipped with probability equal to $0.5$. RandAugment  \cite{Randaugment} follows with $N_{aug}=2$ layers  and magnitude $M_{aug}=7$ (varied using an additive Gaussian noise of a standard deviation equal to $0.5$) and also random erasing augmentation \cite{Random_erase} with probability $0.4$ and $3$ recounts. We diverge from the timm's proposed process only on the resolution of the input \emph{training} images. We reduce the resolution to $160$ (instead of $224$) because on the TPU-v2 of Google Colab the training would require more than three weeks. With the reduced resolution it required approximately $10$ days. The final evaluation of the trained model on the validation set is done using resolution equal to $224$.
    
    As far as the computational cost is concerned, we focus on ImageNet since it is considerably more demanding in terms of computational resources than the CIFAR datasets, 
    and also because for the implementation we use a library dedicated to ImageNet training \cite{timm_tpu}. 
    That way we can directly compare ResNeXt with Coded ResNeXt and focus on the computational impact of Coded ResNeXt's additional steps by minimizing the impact that our implementation may have on the performance.
   
    We would like to clarify that in a ResNeXt block (or Coded ResNeXt block with ratio $r=1$) the last convolutional layer of the block does not have to be implemented as a grouped convolution followed by an aggregation via summation (as it is implied from \cref{fig:arc}).  
    Since for those blocks we do not apply the additional operations of Energy Normalization, coding Loss and dropSubNNs, the grouped convolution with the subsequent aggregation by summation of the subNNs' outputs can be combined into a simple convolutional layer. This is how we implement ResNeXt and blocks of Coded ResNeXt with ratio $r=1$, which also coincides with the way ResNeXt is implemented in the original work \cite{ResNeXt}. 
    
    \begin{table*}[]
        \setlength\extrarowheight{3pt}
      \centering \small
      \begin{tabular}{c||c|c|c|c|}
         & GFlops & \#Params & Throughput & RAM \\
        \toprule
        ResNeXt-50 $(32{\times}4$d) & 2.196 & $25.0\cdot 10^6$ & 378$\mathrm{\frac{samples}{sec}}$ &  12.3GB\\[4pt]
        \hline
        Coded ResNeXt-50 $(32{\times}4$d) & 2.269 & $25.0\cdot 10^6$  &  375$\mathrm{\frac{samples}{sec}}$& 12.7GB\\[4pt]
        \hline
      \end{tabular}
      \caption{Computational cost on ImageNet.}
      \label{tab:Computational_Costs}
    \end{table*}
        
    The throughput and average RAM consumption of \cref{tab:Computational_Costs} have been measured on the second epoch, the reason being that ``generally the first epoch is slow with Pytorch XLA'' \cite{timm_tpu}. 
    For ResNeXt-50 we measure the flops using the library fvcore. 
    For Coded ResNeXt we add to the flops computed for ResNeXt-50 the flops needed for the Energy Normalization step. 
    For a block of $N$ subNNs and output of dimensions $\mathbb{R}^{C\times H \times W}$, the energy normalization requires roughly $3\times N\times C\times H\times W$ flops. The multiplication by $3$ comes from the fact that the energy normalization step first raises in element-wise manner the tensor to the power of $2$, second it takes the mean, and finally it performs an element-wise division with the square root of the total mean energy. We see in \cref{tab:Computational_Costs} that for all metrics, Coded ResNeXt does not introduce any significant additional computational cost when trained on TPU.\footnote{We also tried training on a GPU provided by Google Colab, without relying on timm library. Coded ResNeXt was more than two times slower compared to ResNeXt on that hardware. Nonetheless, the training of both ResNeXt and Coded ResNeXt was faster on TPU, hence we kept TPU as our choice of hardware.}
    
    One epoch of Imagenet on Google Colab TPU-v2 takes roughly $55$ minutes. 
    Due to the platform's constraints, we had to split the training in sessions of 24 hours (i.e., in each session around $25$ epochs were executed\footnote{For the total required time it should be taken into account that around 45 minutes are needed to fetch the dataset from Google Drive to Google Colab per session.}), save the checkpoint at the end of each session, and start the next session from the latest checkpoint stored.
    Due to lack of powerful resources, for ImageNet and with the training setup described above, we tested one additional set of hyperparameters, which was $(\mu,p_{drop})=(4, 0.1)$ (instead of $(2,0.1)$). 
    With $(4,0.1)$ we observed better binary classifiers and early decoders, but lower accuracy. 
    Specifically, the accuracy was $79.81\%$ (instead of $80.24\%$). The  accuracy of the early decoders were  $(4.6\%, 22.0\%,  32.0\%, 37.3\%, 40.0\%,  43.8\%,  55.6\%, 72.5\%, 75.7\%)$ (instead of $(2.38\%, 8.12\%, 12.7\%, 12.8\%, 14.0\%, 17.0\%, 26.5\%, 65.1\%,	73.4\% $)). 
    
    Under a simpler and shorter training procedure that did not require that long period of training (for $150$ epochs, with $(\mu,p_{drop})=(1,0.1)$ and without random erasing data augmentation) we also tested other combinations of coding schemes. The proposed schemes, which at stages (s3, s4) have ratios $(16/32, 8/32)$,  an accuracy of 78.1\% was achieved. We tested as well the two following ones, which all led to worse than $77\%$ accuracy: (i) one where stages  (s2, s3, s4) had ratios $(24/32, 16/32, 8/32)$, (ii) one with $(16/32,8/32,4/32)$, and (iii) finally one where the first 3 blocks of s3 had $r=16/32$, the last 3 blocks of s3 had $r=8/32$, and s4 had $r=4/32$.

\section{Using early decoders to improve confidence calibration }\label{sec:confidence calibration} 
    In  \cref{sec:EarlyDecoder} we showed that the way the coding schemes are designed brings additional useful properties. It is possible to stop the evaluation of the network at an intermediate block, measure the energies of the output of each subNN of that block and predict the class of the input image. 
    Therefore, each block trained to comply with a coding scheme can be used as an ``early decoder'' and produce early predictions. However, the final network's predictions are more accurate. Nonetheless, even when the entire network is evaluated, the early predictions can still be of use. In that case, they can provide a confidence estimation on the correctness of the network's final prediction. As shown in \cref{tab:conditional_accuracies} (see \cref{sec:EarlyDecoder}), the more early decoders agree with the final prediction, the higher are the chances that this prediction is correct. Therefore,  \cref{tab:conditional_accuracies} can be used to obtain confidence levels on the output of a network by verifying how many early decoders agree with the final network's prediction and checking the corresponding line in the table.

    We also show that the predictions of the early decoders can also be employed by confidence calibration methods as a source of extra features. Confidence calibration is the problem of predicting probability estimates that are representative of the true correctness likelihood \cite{ConfidenceCalibration}. 
    The softmax operation, which is commonly used as the final operation of the neural network models for classification, provides for each class the likelihood that the input sample belongs to that class.
    However, it is found in \cite{ConfidenceCalibration} that recent state-of-the-art models provide overconfident predictions. Even when those models predict a wrong class, they erroneously estimate a high  probability that their prediction is actually correct. Calibrating the probabilities associated to the predicted class allows reflecting the true correctness likelihood.
    
    Assume a multi-class classification problem where the input $X\in\mathcal{X}$ and label $Y\in \{1,\cdots,K\}$ are random variables following some ground truth joint distribution $\pi(X,Y)$. Let $g:\mathcal{X}\mapsto \mathbb{R}^K$ be the neural network trained for this problem. For a sample $(x,y)\sim \pi$, it outputs the logits $z=g(x)$ and predicts that the correct label of $x$ is $\hat{y}=\mathop{\mathrm{arg\,max}} z$. Calibrating perfectly this neural network means finding a function $h:\mathbb{R}^K \mapsto  [0,1]$ such that 
    \begin{align} \label{eq:perfect_calibration}
        \mathbb{P}(\hat{Y}=Y|\hat{P}=p )= p, 
    \end{align}
    with $\hat{P} = h(g(X)), \hat{Y}=\mathop{\mathrm{arg\,max}} g(X)$ and $\forall p\in [0,1]$.
    The above probability is taken over the joint distribution $\pi$. An example can help us understand \cref{eq:perfect_calibration}. Assume that we found an $h$ that is a perfect calibrator, which means it satisfies \cref{eq:perfect_calibration}. Suppose that for $p=0.8$, there are exactly $100$ input samples $x \in X$ such that $\hat{p} = h(g(x)) = 0.2$. Then \cref{eq:perfect_calibration} says that the neural network $g(x)$ predicts the correct label, i.e., $\hat{y}=y$, for only $80$ out of those $100$ samples. In other words, the function $h$ correctly estimates that out of all those samples that gave $0.8$ likelihood to be correct, exactly $80\%$ of them were indeed correctly labeled by the neural network $g(x)$.
    
    Since in practice we have datasets with finite number of samples, we have to estimate the performance of the calibrating function $h$ by approximating the probability in \cref{eq:perfect_calibration} \cite{PlattForNN,CalibrationStatistics}. To do that, we break the interval $[0, 1]$ into $M$ interval bins of equal size and let those bins be $I_m=(\frac{m-1}{M},\frac{m}{M}], m\in\{1,\cdots M\}$. If $D$ is a given dataset, then let $B_m = \{(x,y)\in D: h(g(x))\in I_m\}$ be the subset of samples for which the calibrating function $h$ gives that the predicted label is correct with probability belonging to the interval $I_m$. An unbiased estimator of $\mathbb{P}(\hat{Y}=Y|\hat{P} \in I_m)$ is 
    \begin{align}
        \mathrm{acc}(B_m) = \frac{1}{|B_m|}\sum_{(x,y)\in B_m}\mathds{1}(\hat{y}=y)
    \end{align}
    with $\mathds{1}$ being the indicator function. The average confidence within bin $B_m$ is defined as
    \begin{align}
        \mathrm{conf}(B_m) = \frac{1}{|B_m|}\sum_{(x,y)\in B_m}\hat{p}
    \end{align}
    with $\hat{p}=h(g(x))$. 
    The function $\mathrm{acc}(\cdot)$ approximates the left-hand side of \cref{eq:perfect_calibration}, while $\mathrm{conf}(\cdot)$ does it for the right-hand side. Therefore, the closer those two values are, the better the calibration function $h$ is. 
    
    Two metrics used to evaluate the calibration are \cite{CalibrationStatistics}:
    \begin{itemize}[leftmargin = *]
        \item Expected Calibration Error (ECE), which approximates $\mathbb{E}[|\mathbb{P}(\hat{Y}=Y|\hat{P}=p ) - p|]$ as $$\mathrm{ECE} = \sum_{m=1}^M \frac{|B_m|}{|B|}|\mathrm{acc}(B_m)-\mathrm{conf}(B_m)|, $$
        \item Maximum Calibration Error (MCE), which approximates $\underset{p\in[0,1]}{\max}|\mathbb{P}(\hat{Y}=Y|\hat{P}=p ) - p|$ as $$\mathrm{MCE} = \underset{m\in{1,\cdots,M\}}}{\max}|\mathrm{acc}(B_m)-\mathrm{conf}(B_m)|. $$
    \end{itemize}
    ECE measures on average how well calibrated the function $h$ is and MCE provides the maximum error of the predicted likelihoods. 
    MCE can be a valuable metric in high-risk applications where good reliability guarantees should be provided even in the worst-case scenario.

    Surprisingly, in \cite{ConfidenceCalibration} it was found that the calibration function that works the  best is simply a softmax function with a single parameter $T$ regulating the scale (also known as ``temperature''). This is an extension of Platt scaling \cite{Platt,PlattForNN} and the calibration function used is 
    \begin{align}\label{eq:Calibration_Platt}
        h_T(z)_k = \frac{e^{z_k/T}}{\sum_{j=1}^K e^{z_j/T}}, k\in\{1,\cdots,K\}
    \end{align}
    where $z\in\mathbb{R}^K$ is a vector (representing the logits produced by the model, i.e., $z=g(x)$) and $z_k$ is the $k$-th element of that vector.
    The value of the parameter $T$ is chosen by minimizing the negative log-likelihood on a hold-out validation set \cite{ConfidenceCalibration}. The minimization is performed using gradient descent updating only the parameter $T$. We emphasize that since the parameters of the network are kept fixed, the network predictions have not changed. $h_T$ calibrates the likelihood probabilities associated to the class predicted by the network.
    
    We propose now a simple way to take into account the predictions of the early decoders. Given an input sample, suppose that $\hat{e_i}, i\in\{1,\cdots,N_{dec}\}$ is the prediction of the $i$-th early decoder, $N_{dec}$ is the number of early decoders, and $\hat{y}$ the final prediction. Our proposed architecture for CIFAR has $N_{dec}=6$ and for ImageNet $N_{dec}=9$. We intend to adjust the parameterized function $h_T$ by adding another $N_{dec}$ parameters, denoted $\tau_i$, which account for the predictions of the early decoders. Specifically,  \textit{the calibration function} we employ is
    \begin{align}\label{eq:Calibration_with_earlyDecoders}
        h_{aug}(z)_k = \frac{e^{z_k/T_{aug}}}{\sum_{j=1}^K e^{z_j/T_{aug}}}, \textrm{ with } T_{aug}= T\left(1 + \sum_{i=1}^{N_{dec}}\mathds{1}(\hat{e_i}=\hat{y})\tau_i\right)  \textrm{ and } k\in\{1,\cdots,K\},
    \end{align}
    which has $N_{dec}+1$ parameters. If the $i$-th early decoder agrees with the final prediction $\hat{y}$ then the temperature is adjusted by adding the term $\tau_i$. The parameters are optimized in the same way as for $h_T$, using gradient descent with respect to the negative log-likelihood. We initialize all $\tau_i$ to be zero and $T$ to be one. The learning rate is $0.1$ and we perform $600$ iterations. 
    
    We start the experimental procedure by randomly splitting the validation set of each dataset into two sets. The first contains a quarter of the total samples and the second the rest. The splitting is performed in a way such that within the set, all classes have approximately the same number of samples. We use the first set to train the parameters of $h_T$ and $h_{aug}$. We test their performance on the second set by estimating the ECE and the MCE, using $M = 15$ bins as in \cite{ConfidenceCalibration}. We repeat the procedure $40$ times more, each time with a different random split of the validation dataset. 
    The results are shown in \cref{tab:ECE_MCE_quarter}. 
    In \cref{tab:ECE_MCE_half} we conduct the same experiment but now we split the validation set into two sets of equal size. 
    Across all datasets and for both splitting ratios we see that $h_{aug}$ improves the MCE with almost no impact on ECE with respect to $h_T$.

    \begin{table}
        \begin{subtable}[c]{0.5\textwidth}
            \centering
            \begin{tabular}{c||c|c|}
              &  \thead{ $h_T$ of \cref{eq:Calibration_Platt}\\from \cite{ConfidenceCalibration}} & \thead{ $h_{aug}$ of \cref{eq:Calibration_with_earlyDecoders}\\Ours} \\
            \hline
             CIFAR-10  & (0.70\%, 24.35\% ) & ( 0.70\%, \textbf{24.14}\%)  \\
            \hline
             CIFAR-100 & (1.75\%, 10.28\%) & ( \textbf{1.71}\%,\textbf{ 9.31}\% )\\
            \hline
             ImageNet &  (\textbf{2.98}\%, 8.32\%) & (3.00\%, \textbf{7.70}\%) \\
            \hline
            \end{tabular}
        \subcaption{Split set into: train set $25\%$, test set $75\%$.}
        \label{tab:ECE_MCE_quarter}
        \end{subtable}
        \begin{subtable}[r]{0.57\textwidth}
            \centering
            \begin{tabular}{|c|c|}
              \thead{ $h_T$ of \cref{eq:Calibration_Platt}\\from \cite{ConfidenceCalibration}} & \thead{ $h_{aug}$ of \cref{eq:Calibration_with_earlyDecoders}\\Ours} \\
            \hline
              (0.80\%, 24.83\% ) & ( 0.80\%, \textbf{23.64}\%)  \\
            \hline
              (1.82\%, 11.37\%) & ( \textbf{1.79}\%,\textbf{10.44}\% )\\
            \hline
             (\textbf{3.01}\%, 8.56\%) & (3.05\%, \textbf{8.10}\%) \\
            \hline
            \end{tabular}
        \subcaption{train set 50\%, test set 50\%}
        \label{tab:ECE_MCE_half}
        \end{subtable}
    \caption{The performance in terms of (ECE(\%), MCE(\%)) of the two calibration methods using temperature scaling. The first one $h_T$ is proposed in \cite{ConfidenceCalibration} and has only one parameter. The second $h_{aug}$ has additionally one parameter per early decoder and takes into account the prediction of the early decoders.}
    \end{table}

\section{Ablation Study on Coding Loss and dropSubNN}\label{sec:ablation}

    \begin{figure*}[h!]
      \centering
      \begin{subfigure}{0.49\linewidth}
      \centering
       \includegraphics[width=1.0\columnwidth]{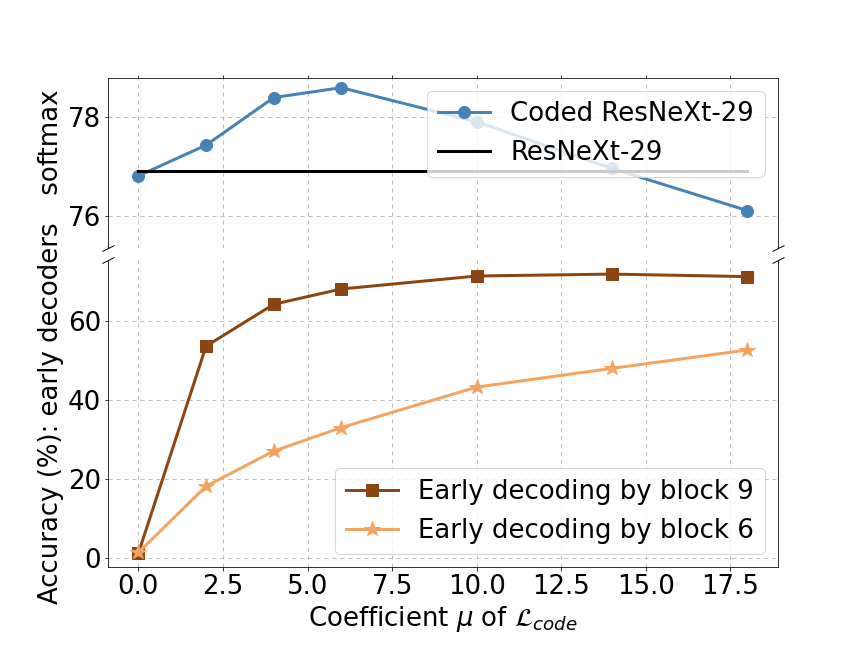}
       \caption{Impact of coefficient $\mu$ of the coding loss on {CIFAR-100}.}
       \label{fig:ablation_tradeOff_losses}
      \end{subfigure}
      \hfill
      \begin{subfigure}{0.49\linewidth}
      \centering
       \includegraphics[width=1.0\columnwidth]{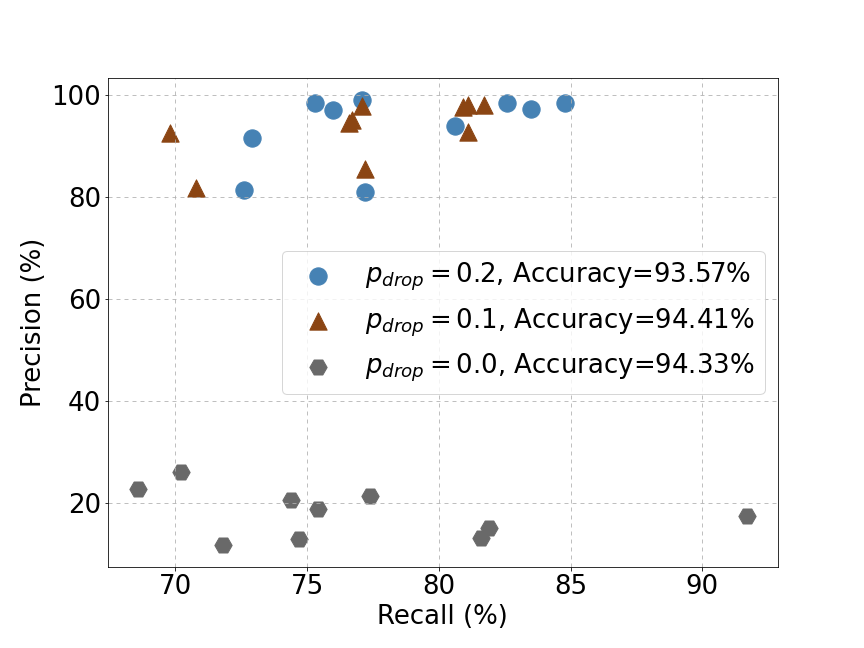}
       \caption{Impact of the probability $p_{drop}$ on the performance of the binary classifiers. }
       \label{fig:ablation_dropSubNN}
      \end{subfigure}
      \caption{Ablation study on the two hyperparameters introduced in this paper, i.e., the coefficient $\mu$ balancing the losses in \cref{eq:total_loss} and the probability $p_{drop}$ of dropping subNNs.}
      \label{fig:ablation}
    \end{figure*}
    
    In this section, we study the effect of the two hyperparameters introduced in the paper, namely the coefficient $\mu$ balancing the losses in \cref{eq:total_loss} and the probability $p_{drop}$ of dropping subNNs.
    
    In \cref{fig:ablation_tradeOff_losses} we show that increasing $\mu$ and so forcing more the energies of the subNNs to comply with the coding scheme, is at first beneficial to the overall performance until a certain point (see blue line). 
    Past this point (in \cref{fig:ablation_tradeOff_losses} it is around $\mu=6.0$), forcing the subNNs to output a signal of a specific energy value provides only small diminishing gains on the early decoders (see increasing trend of brown and beige lines). Furthermore, it disturbs the entire classification process, and the final accuracy of the whole network's predictions starts declining.
    
    The second experiment concerns the dropSubNN and its hyperparameter $p_{drop}$. Dropping randomly some subNNs during training inhibits their ``co-adaptation''\cite{dropout}, as they learn not to depend on the others and to perform well even in the absence of some of them. \Cref{fig:ablation_dropSubNN} shows that the dropSubNN is essential for the good performance of the binary classifiers. A small value of $p_{drop}=0.1$ can greatly boost their performance and even slightly improve the overall accuracy. Further increasing it to $p_{drop}=0.2$ degrades the accuracy without improving much the binary classifiers.

\section{Limitations} \label{sec:limitations}
    A limitation regarding the proposed idea is that it cannot be applied to any type of multi-branch architecture. As we explain in \cref{sec:whyResNeXt} the method works for ResNeXt because the output of all branches of every block is conveniently aggregated by summation. 
    This avoids having the blocks relying on the output of the inactive subNNs to learn the classification, which would then prevent the extraction of the binary classifiers by just keeping the active subNNs. Instead, since (i) the blocks receive the aggregated outputs of the previous block, and (ii) the outputs of the inactive subNNs are pushed to zero by the coding loss, during training the blocks are forced to solely rely on the aggregated signal provided by the active subNNs.
    Interestingly, there are other architectures that have a structure similar to that of ResNeXt and where our idea could also be applied. An example is MobileNetV2 \cite{R3_MobilenetV2}, which uses a type of block called MBConv (a block that later used also by EfficientNet \cite{R3_efficientnet}). To see how our idea could be applied to this kind of blocks, let us define the following neural network module in PyTorch:
    \floatname{algorithm}{}
    \begin{algorithm}
        \renewcommand{\thealgorithm}{}
        \caption{Pytorch sequential code}\label{alg:Module code}
        \begin{algorithmic}[1]
        \Require channels\_in, channels\_mid, channels\_out, $N$, $s$
        \State module = nn.Sequential(
        \State \hspace{0.3cm} nn.Conv2d(channels\_in, channels\_mid, kernel\_size=1),
        \State \hspace{0.3cm} nn.BatchNorm2d(channels\_mid),
        \State \hspace{0.3cm} nn.ReLU(),
        \State \hspace{0.3cm} nn.Conv2d(channels\_mid, channels\_mid, kernel\_size=3, padding=1, groups=$N$, stride=$s$),
        \State \hspace{0.3cm} nn.BatchNorm2d(channels\_mid),
        \State \hspace{0.3cm} nn.ReLU(),
        \State \hspace{0.3cm} nn.Conv2d(channels\_mid, channels\_out, kernel\_size=1),
        \State \hspace{0.3cm} nn.BatchNorm2d(channels\_out) 
        \State \hspace{0.3cm} )
        \end{algorithmic}
    \end{algorithm}
    
    A typical ResNeXt block is formed by simply adding a residual connection to the above module, i.e., module($x$)$+x$ where $x$ is the input of the block.\footnote{We assumed that channels\_in = channels\_out, and therefore there is no expansions of the number of channels which usually happens in the beginning of each stage.} This ResNeXt block has $N$ branches/subNNs. The MBConv uses an almost identical module, with the slight differences that (i) the activation function ReLU6($\cdot$) is used instead of ReLU($\cdot$), (ii) channels\_mid takes a larger value than channels\_out, whereas in ResNeXT it takes a smaller one, and (iii) it forces $N=$ channels\_mid so the second convolution is depth-wise. Overall, the MBConv can be seen as a type of ResNeXt block but with the number of subNNs equal to the channels\_mid. We therefore expect a ``Coded-MobileNetV2'' to behave the same as Coded-ResNeXt and exhibit the same properties.
    
    Another limitation of our work is given by the trade-off between the specialization of the subNNs and the performance of the entire neural network model.
    This relationship is clearly depicted in \cref{fig:ablation_tradeOff_losses}. To force the subNNs to specialize in a specific set of classes we introduced the ``coding loss'' $\mathcal{L}_{code}$. The balance between this loss and the cross entropy loss (i.e., $\mathcal{L}_{class}$) is regulated with the hyperparameter $\mu$. As $\mu$ increases, more emphasis is given on the subNNs activated according to the coding scheme. 
    As shown in \cref{fig:ablation_tradeOff_losses}, increasing $\mu$ at first helps the total performance (accuracy) of the entire network. 
    However, after a certain point, increasing the specialization of the subNNs comes at the expense of the performance of the entire network.
    
    We came across this trade-off also when designing the ratios $r$ of the coding schemes. Ideally, we would like the ratios to be as small as possible. This would mean that each subNN would be assigned to specialize to smaller set of classes. Consequently, the information paths of the classes would be more disentangled, in the sense that the paths associated to two different classes would have less shared parameters. Moreover, the extracted binary classes would have even less parameters. Unfortunately, we could not decrease more the ratios without negatively impacting the total model's accuracy. For example, in one of our initial experiments (with fewer epochs)\footnote{(150 epochs, with $(\mu, p_{drop}) = (1, 0.1)$ and without random erasing data augmentation)} on ImageNet, we found that if we used the proposed ratios of $(32/32, 16/32, 8/32)$  in stages $(c3, c4,c5)$ we get an accuracy of $78.1\%$, but trying to increase the subNNs' specialization by setting the ratios to $(16/32, 8/32, 4/32)$ drops the accuracy to $76.6\%$.  
    
    Nonetheless, we are confident that there exist ways in which a more drastic specialization of the subNNs is achieved without compromising the performance of the entire network.
    One alternative that was the subject of some of our tests was zeroing the gradients during training\footnote{This can be implemented on PyTorch using the detach function.}, which are directed to the subNNs that according to the coding scheme should remain inactive. 
    That way, the subNNs are updated only by gradients coming from samples of the subset of classes that the coding scheme had assigned to them. 
    Those experiments were performed in a network whose architecture was identical to Coded ResNeXt with the sole difference that the blocks had no skip/residual connections. 
    In that architecture, replacing after Energy Normalization the $\mathcal{L}_{code}$ with an operation that zeroes the gradients according to the coding schemes yielded excellent specialization without degrading the overall network's performance. 
    Unfortunately, this idea did not work well when we added the skip connections. The accuracy of the network was significantly lower when the architecture had skip connections and the gradients of inactive subNNs were zeroed during training.

\section{Additional information for binary classifiers and activation of subNNs}\label{sec:BinaryClassifiersAdditionalPlots}
    
    In this section, we provide additional details regarding the binary classifiers that can be extracted after training a Coded ResNeXt, the activation distributions of the subNNs, and their specialization. 
    
    \subsection{Activation distribution of subNNs}\label{subsec:Activation Distribution}
\begin{figure*}[ht]
      \centering
      \begin{subfigure}{0.34\linewidth}
        \includegraphics[width=1.0\columnwidth]{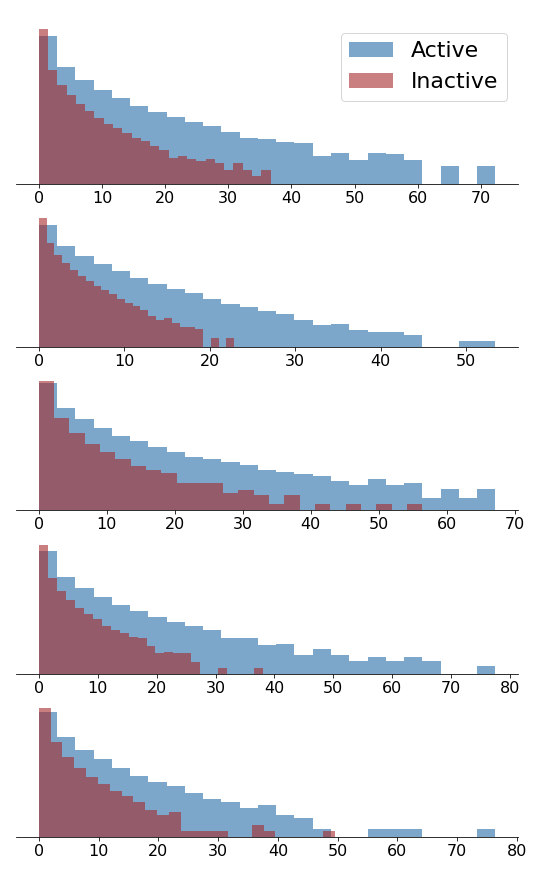}
       \caption{Distribution of the absolute value of the output of the first 5 subNNs from the second block of stage s3 on the architecture for CIFAR-10.}
       \label{fig:1}
      \end{subfigure}
      \hfill
      \begin{subfigure}{0.34\linewidth}
        \includegraphics[width=1.0\columnwidth]{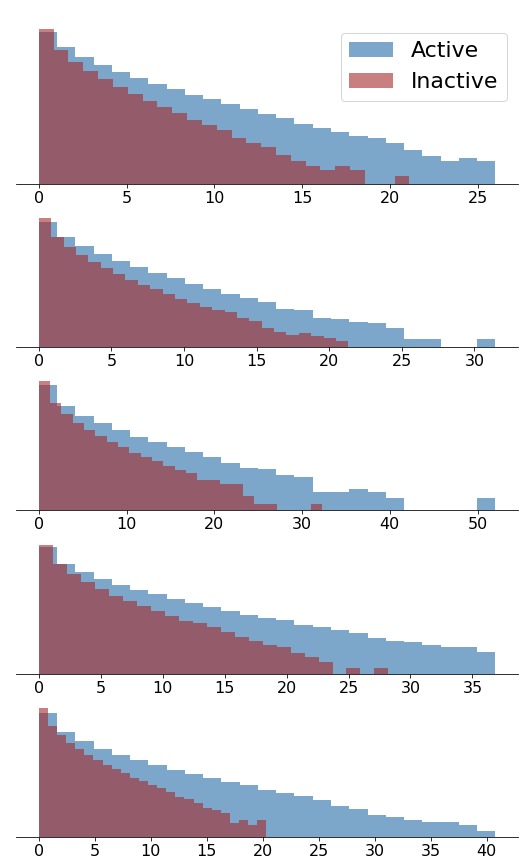}
       \caption{Distribution of the absolute value of the output of the first 5 subNNs from the second block of stage s3 on the architecture for CIFAR-100.}
       \label{fig:2}
      \end{subfigure}
      \hfill
      \caption{Output distribution of active versus inactive subNNs }
      \label{fig:Activation_distributions}
    \end{figure*}

    In this subsection we investigate the distribution of the output signal of the subNNs. A sample that passes through a subNN can either belong to the set of classes for which the subNN has to activate, or to the set for which it has to stay inactive. We would like to see how the subNNs respond in these two scenarios by plotting the distribution of the absolute value of the output in each case. In \cref{fig:Activation_distributions} we plot for CIFAR-10 and CIFAR-100 those distributions for the first 5 subNNs of the second block of stage s3. We show in red how those subNNs react when the sample belongs to the set of classes for which the subNN has to stay inactive according to the coding scheme, and with blue when they have to activate. Since the distributions were very skewed with many values around zero, we show the y-axis in log scale to better appreciate their shape in the complete range. We see that the distribution of the active subNNs has a bigger tail, which confirms that the subNNs output higher values when receiving a sample of their assigned classes.
    
    \newpage
    \subsection{Additional details on Binary Classifiers}\label{subsec:Additional plots BC}

        \begin{figure}[t]
          \centering
          \begin{subfigure}{0.32\linewidth}
             \includegraphics[trim=25mm 70mm 16mm 72mm, clip,width=1.0\columnwidth]{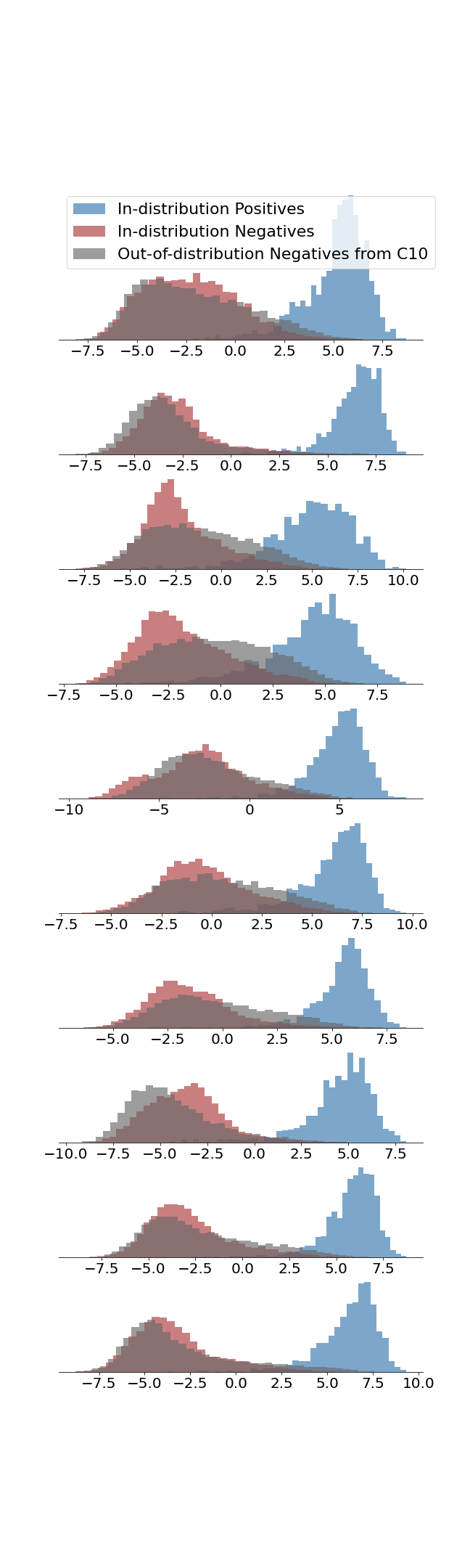}
           \caption{All Binary Classifiers extracted from Coded ResNeXt-29 $(10\times 11$d) trained on CIFAR-10.\\ $\quad$}
           \label{fig:BinaryClassifier_first10Classes_CIFAR10}
          \end{subfigure}
          \hfill
          \begin{subfigure}{0.32\linewidth}
            \includegraphics[trim=25mm 70mm 16mm 72mm, clip,width=1.0\columnwidth]{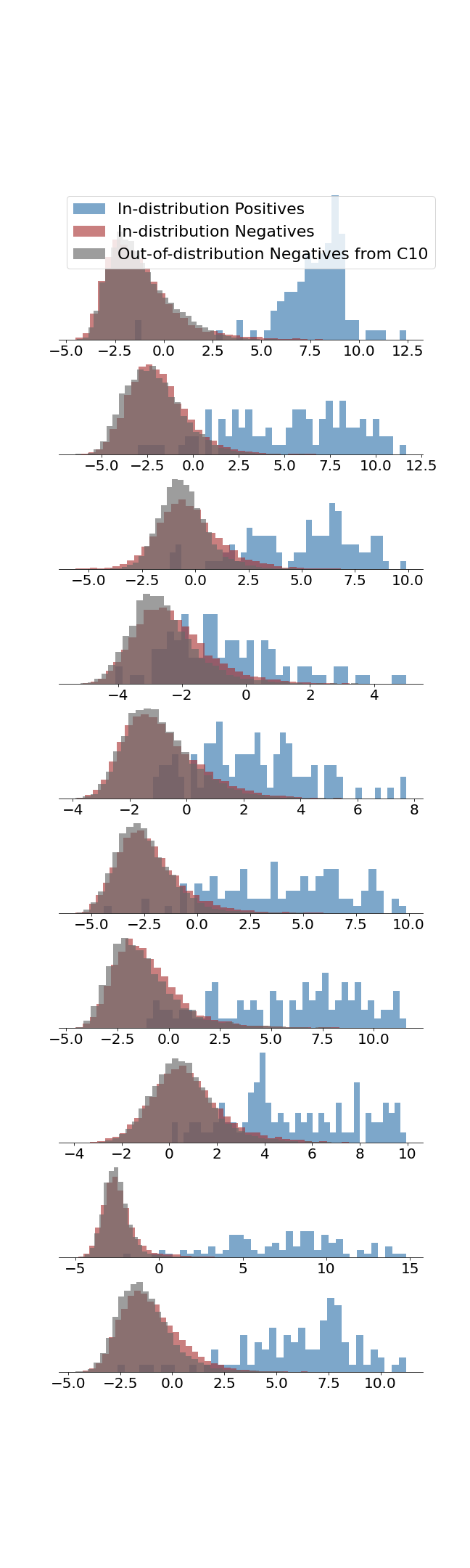}
           \caption{Binary Classifiers for the first 10 classes extracted from Coded ResNeXt-29 $(20\times 6$d) trained on CIFAR-100.}
           \label{fig:BinaryClassifier_first10Classes_CIFAR100}
          \end{subfigure}
          \hfill
          \begin{subfigure}{0.32\linewidth}
            \includegraphics[trim=25mm 70mm 16mm 72mm, clip,width=1.0\columnwidth]{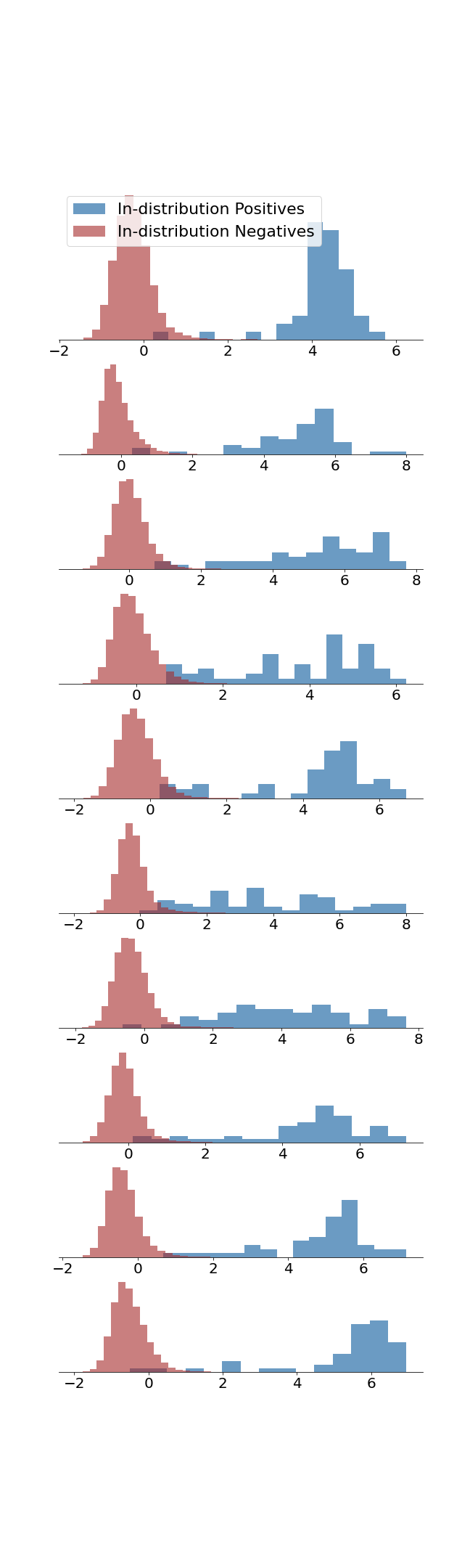}
           \caption{Binary Classifiers for the first 10 classes extracted from Coded ResNeXt-50 $(32\times 4$d) trained on ImageNet.}
           \label{fig:BinaryClassifier_first10Classes_ImageNet}
          \end{subfigure}
          \caption{Distribution of the output (logit) of binary classifiers}
          \label{fig:BinaryClassifier_for_10_classes_per_dataset}
        \end{figure}
    
        \begin{wrapfigure}{r}{0.48\textwidth}
        \vspace{-1cm}
          \centering
           \includegraphics[trim=0mm 10mm 10mm 0mm, clip, width=0.49\textwidth]{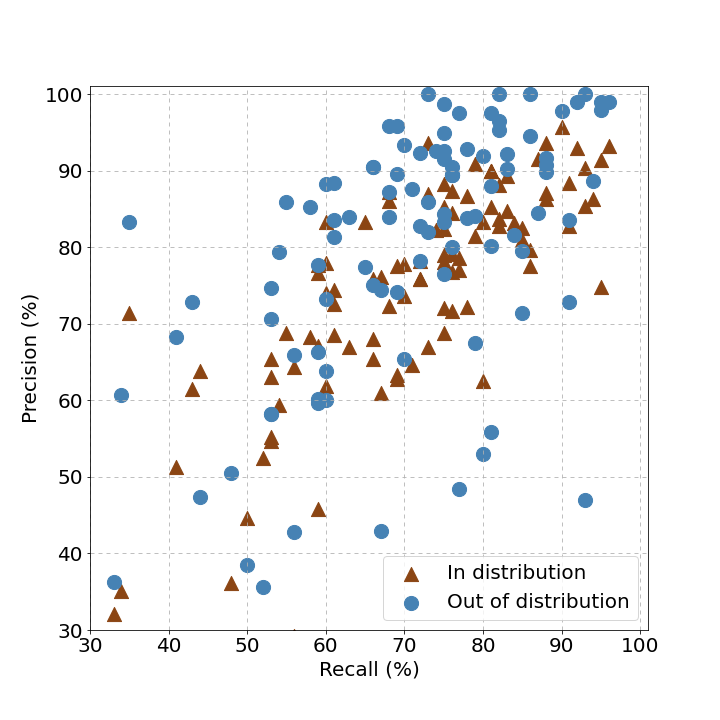}
           \caption{Precision-Recall on CIFAR-100.}
        \vspace{-0.2cm}
           \label{fig:PrecisionRecall_CIFAR100}
        \end{wrapfigure}
        
        For convenience, we provide here the definitions of Precision and Recall. True positives $TP$ (resp. true negatives $TN$) represents the number of positive (resp. negative) samples that the binary classifier correctly predicts as positives (resp. negatives). False positives $FP$ (resp. false negatives $FN$) represents the number of negative (resp. positive) samples that the binary classifier erroneously predicts as positives (resp. negatives). The precision and recall are defined as follows 
        $$ Precision=\frac{TP}{TP+FP},\; Recall=\frac{TP}{TP+FN}.$$
        Precision measures out of predicted positives (i.e., $TP+FP$) how many of those predictions are correct. Recall measures out of all actual positives (i.e., $TP+FN$) how many were found by the binary classifier.
        
        In \cref{fig:BinaryClassifier_for_10_classes_per_dataset} we give the output distributions of the binary classifiers for the first 10 classes of CIFAR-10/100 and ImageNet when fed with an input of in-distribution positive/negative samples, and out-of-distribution negative samples. The first row coincides with \cref{fig:BinaryClassifiers}. We observe that depending on the threshold value a binary classifier uses (above which the classifier predicts positive and below negative), different values of precision and recall can be attained. Increasing the threshold value gives fewer false positives $FP$, but unfortunately more false negatives $FN$ as well, so increasing the threshold improves the precision but degrades the recall. 
        Therefore, high recall can be exchanged for high precision by increasing the threshold, or the opposite by decreasing it.

        Regarding plots \cref{fig:PrecisionRecall_CIFAR10}, \cref{fig:PrecisionRecall_CIFAR100} and \cref{fig:PrecisionRecall_ImageNet}, which depict the precision and recall achieved by the binary classifiers trained on CIFAR-10, CIFAR-100, and ImageNet, respectively, we would like to make some remarks.\footnote{In \cref{fig:PrecisionRecall_CIFAR100}, which corresponds to CIFAR-100, we followed the same procedure as in \cref{fig:BinaryClassifier_ImageNet} for ImageNet. Specifically, we randomly sample from the set of negatives a subset of size 9 times bigger than the size of positives. We use that subset of negatives to evaluate the precision and recall of the binary classifier.} First, when testing a binary classifier in the case of either in-distribution or out-of-distribution negatives, the set of samples of the validation set serving as positives remains the same. Since by definition recall depends only on this set of positives, its value remains unaltered in both testing cases. Second, in the case of ImageNet, we only have 50 samples of the validation set serving as positives for each binary classifier. Therefore, the true positives can only take integer values from 0 to 50 and $Recall\in\{ 0, \frac{1}{50}, \frac{2}{50}, \cdots, \frac{50}{50}\}$. This is the reason why in \cref{fig:PrecisionRecall_ImageNet}, the points appear to follow some kind of grid and are aligned in specific vertical lines. 
        
        Finally, and quite interestingly, we see that the binary classifiers trained on CIFAR-100 perform at the same level no matter whether the negatives come from in-distribution or out-of-distribution. This is in contrast to the binary classifiers trained on CIFAR-10, where out-of-distribution negatives clearly decrease the precision \cref{fig:PrecisionRecall_CIFAR10,fig:BinaryClassifier_for_10_classes_per_dataset,fig:PrecisionRecall_CIFAR100}). Both CIFAR-10 and CIFAR-100 datasets have the same number of training samples (50000), but CIFAR-10 has 5000 samples per class and CIFAR-100 has 500 per class. Therefore, the binary classifiers of CIFAR-100 have been trained on $\frac{5000}{500}=10$ times fewer positives but have ``seen'' negatives from $\frac{100-1}{10-1}=11$ times more classes. 
        The observation that CIFAR-100 binary classifiers perform significantly better than the ones of CIFAR-10 with out-of-distribution negatives, agrees with our intuition that a model should work better for out-of-distribution data if it has been trained with a high variety of classes where the number of samples per class is small, than if it is trained with the same amount of samples but fewer classes and more examples per class.
        

    \subsection{More experiments on removing subNNs}\label{subsec:Additional plots on removing subNNs}
\begin{figure*}[ht]
          \centering
          \begin{subfigure}{0.322\linewidth}
            \includegraphics[width=1.0\columnwidth]{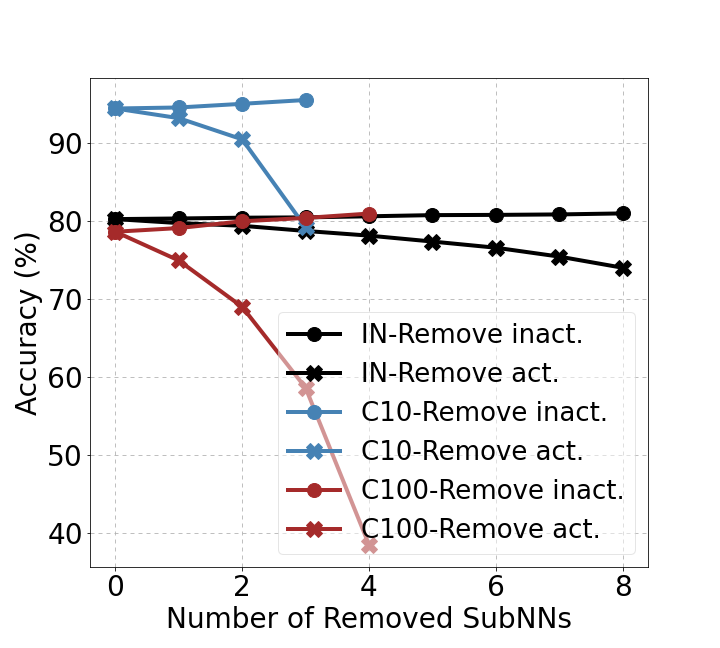}
           \caption{Removing subNNs from the last block.}
           \label{fig:RemovingSubNNs_LastBlock}
          \end{subfigure}
          \hfill
          \begin{subfigure}{0.34\linewidth}
            \includegraphics[trim=0mm 0mm 20mm 0mm, clip,width=1.0\columnwidth]{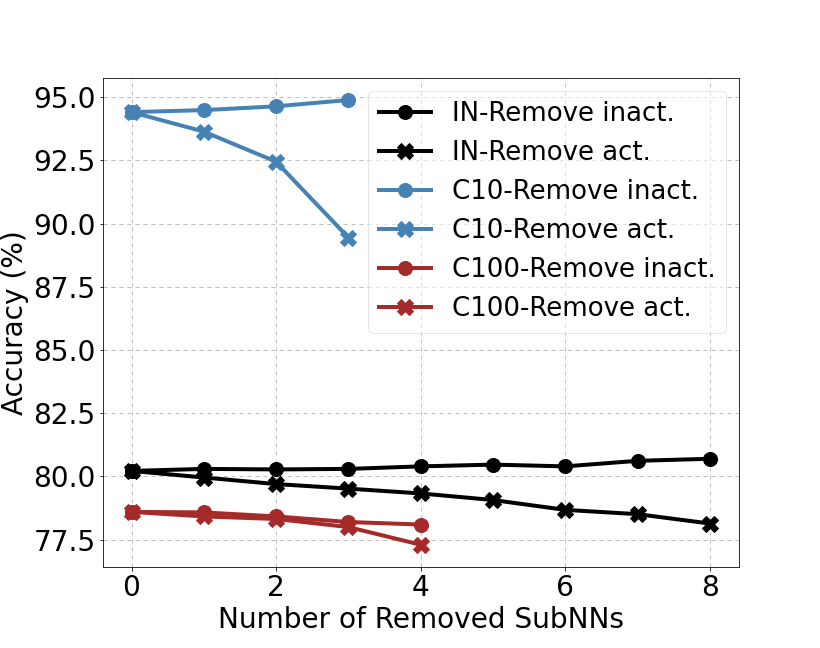}
           \caption{Removing subNNs from the third to last block.}
           \label{fig:RemovingSubNNs_Third2Last_Block}
          \end{subfigure}
          \hfill
          \begin{subfigure}{0.319\linewidth}
            \includegraphics[width=1.0\columnwidth]{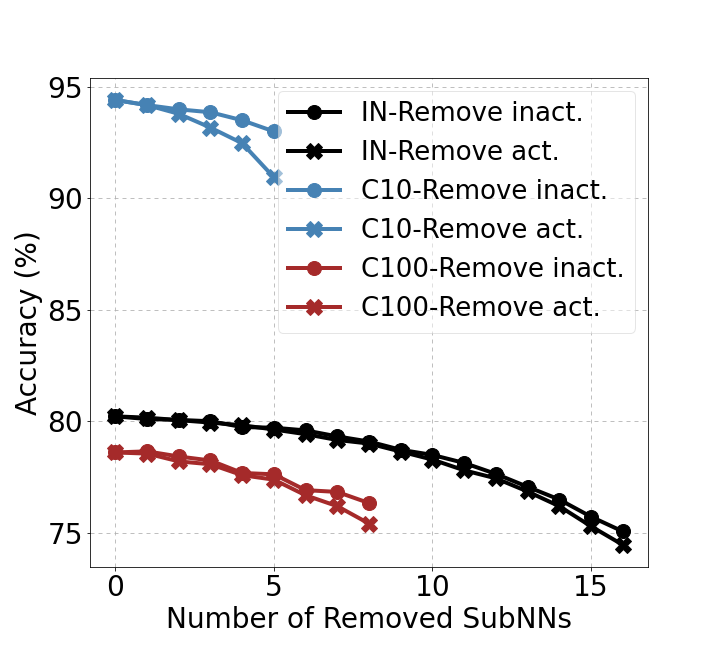}
           \caption{Removing subNNs from the fifth to last block.}
           \label{fig:RemovingSubNNs_Fifth2Last_Block}
          \end{subfigure}
          \caption{Performance when removing active versus inactive subNNs from a specific block. }
          \label{fig:RemovingSubNNsFromManyBlocks}
        \end{figure*}

        In this subsection, we repeat the experiment described in \cref{sec:Experiments_setup} for three more blocks. 
        We remind the reader that in that experiment we pick a block $l$ from which we randomly remove subNNs in two ways: given the class of the input image sampled from the validation set, the first way randomly removes $k{\leq} N_{act,l}$ subNNs from the set of active for that class subNNs. 
        The second way randomly removes $k{\leq} N-N_{act,l}$ subNNs from the (complementary) set of inactive subNNs for that class. 
        We see that the trend depicted in \cref{fig:RemovingSubNNs_GivenBlock}, where the performance drops more when active subNNs are removed from a block than when the inactive ones are removed,
        is also maintained for the blocks in \cref{fig:RemovingSubNNsFromManyBlocks}. 
        
        However, we see that this trend becomes less visible the earlier is the block in which we perform the experiment. For example in \cref{fig:RemovingSubNNs_Fifth2Last_Block} we see that removing either active or inactive subNNs leads to small differences. 
        Removing active subNNs from the earlier blocks has much less negative effect than from the blocks deeper in the architecture. 
        We believe that the reason is that the subNNs in the early blocks produce low level features, which most of the times are useful in general for any class.
        Therefore, it is challenging to force those subNNs to produce low level features that are exclusively useful to a specific subset of classes. 
        This difficulty is more pronounced in this study, where we design the coding schemes without taking into account the semantics of each class. 
        Stated differently, it is hard, given a set of classes and without looking at the semantic similarities between the classes, to split the set into two subsets and push subNNs to produce low level features that are useful only for one of those subsets and useless for the other.   
        Nonetheless, we expect that increasing $\mu$, which could push the subNNs towards higher specialization, would lead to having the performance degradation due to removing active subNNs observed in the latter blocks also in early blocks.
        

\end{document}